\documentclass{article}
\usepackage[preprint]{neurips_2026}

\usepackage[utf8]{inputenc}
\usepackage[T1]{fontenc}
\usepackage{hyperref}
\usepackage{url}
\usepackage{booktabs}
\usepackage{amsfonts}
\usepackage{amsmath,amssymb}
\usepackage{nicefrac}
\usepackage{microtype}
\usepackage{xcolor}
\usepackage{graphicx}
\usepackage{caption}
\usepackage{subcaption}
\usepackage{float}
\usepackage{dblfloatfix}
\graphicspath{{figures/}}
\usepackage{algorithm}
\usepackage[noend]{algpseudocode}
\usepackage[table]{xcolor}
\usepackage{multirow}
\usepackage{enumitem}
\usepackage{bbm}

\title{Counterfactual Fragility Certificates: Exposing High-Confidence Brittleness under Structured Evidence Failure}

\author{
Filippo Cenacchi, Longbing Cao, and Runze Yang\\
Macquarie University, Sydney, Australia\\
\texttt{filippo.cenacchi@mq.edu.au, longbing.cao@mq.edu.au, runze.yang@hdr.mq.edu.au}
}

\begin{document}

\maketitle

\begin{abstract}
High test accuracy and good aggregate calibration do not show whether an individual prediction is structurally supported by its evidence. In tabular decision systems, failures often occur when a feature family becomes unavailable, delayed, noisy, stale, or low-trust while the model remains highly confident. Existing calibration, uncertainty, selective-prediction, explanation, and perturbation methods provide scalar scores or attribution maps, but not a recomputable audit object answering: \emph{under a declared evidence-failure protocol, what trajectory makes this prediction lose support?} We introduce \emph{Counterfactual Fragility Certificates} (CFC), a model-agnostic protocol-level audit certificate---not a formal robustness certificate---that maps each prediction into an ordered evidence-failure trajectory summarized by a greedy flip budget, normalized margin-collapse area, degradation thresholds, and fragility dominance score. CFC is a deterministic witness under fixed grouping, baseline, stress operators, severity grid, and audit depth. Across seven tabular benchmarks and strong linear, tree-based, boosting, and neural baselines, CFC-FDS identifies independently brittle high-confidence cases with 0.915 AUROC, improving over the strongest non-certificate score by +0.405. The advantage persists against perturbation, permutation-importance, group-SHAP, baseline-choice, seed-variance, budgeted-review, and naturalistic field-unavailability checks. Under a 20\% review budget, CFC-FDS captures 88.9\% of brittle high-confidence cases, compared with 31.8--37.4\% for confidence and energy scores. We additionally evaluate fragility-aware regularization and brittleness-aware temperature correction as secondary uses. CFC provides a concrete reliability framework for exposing high-confidence brittleness that ordinary score-centric evaluation misses.
\end{abstract}

\section{Introduction}
Average-case evaluation still dominates machine learning reporting, but deployment failures often concentrate in predictions that look strong until their supporting evidence is stressed. A classifier can post high AUROC, macro-F1, and negative log-likelihood while relying on a dangerously narrow support set for some cases. Calibration work shows that modern predictors can be sharply overconfident and that post-hoc methods such as temperature scaling, Bayesian binning, and Dirichlet calibration can improve probability semantics without changing the decision rule \citep{guo2017calibration,naeini2015obtaining,kull2019beyond,kumar2018trainable,minderer2021revisiting,wang2021rethinking,dheur2023probcal}. Selective prediction adds abstention and confidence-based filtering when risk is high \citep{geifman2017selective,geifman2019selectivenet,liu2019deep,corbiere2019addressing,moon2020confidence,traub2024overcoming}. Yet these methods still reduce reliability mainly to scalar confidence and do not measure whether predictions remain stable under structured evidence degradation. This matters in tabular systems, where failures often arise from missing, delayed, stale, or low-quality feature groups rather than adversarial noise. Recent tabular benchmarks show that tree ensembles and foundation-style models remain difficult to dominate \citep{gorishniy2021revisiting,grinsztajn2022why,hollmann2025accurate}; robustness studies show that realistic shift and adversarial stress remain unresolved \citep{gardner2023benchmarking,simonetto2024tabularbench}; and selective-classification studies show that confidence-based rejection can miss undetected-error risk \citep{fisch2022calibrated,ding2023topambiguity,traub2024overcoming,wu2024confidenceaware,zhu2022rethinking,tomani2023densityaware}. The missing artifact is a standard, recomputable object that quantifies how rapidly a prediction breaks when semantically meaningful feature groups are weakened or removed.

\begin{figure}[H]
    \centering
    \vspace{-0.6em}
    \includegraphics[width=0.75\textwidth]{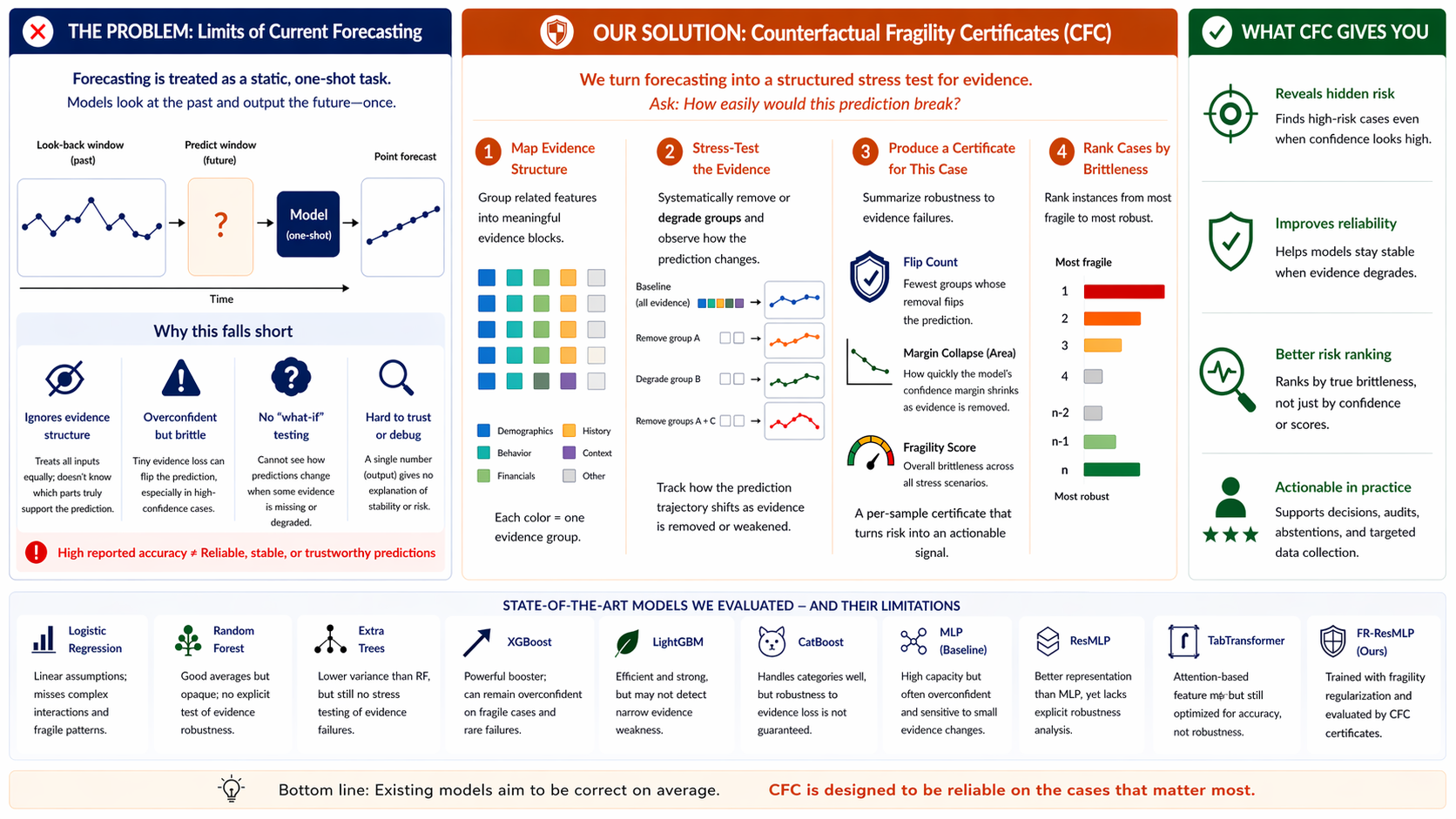}
    \caption{\textbf{CFC overview.}
CFC stress-tests feature groups under evidence loss and returns case-level certificates for brittleness-aware ranking and decision support.}
    \label{fig:teaser}
    \vspace{-1.1em}
\end{figure}

Figure~\ref{fig:teaser} frames this gap; we address it with \emph{Counterfactual Fragility Certificates} (CFC), a per-sample, model-agnostic audit object computed through controlled forward passes over a trained model and grouped preprocessed features. Unlike confidence, calibration error, attribution, one-step perturbation importance, or counterfactual recourse, CFC records a declared \emph{failure path}: ordered evidence states, first prediction flip, margin-collapse area before or without a flip, and the severity at which partial degradation becomes decision-changing. The unit of reliability analysis therefore changes from a scalar score to an operational question: \emph{under this stated evidence-failure protocol, what trajectory makes the prediction lose support?} The certificate is deterministic and inspectable, but intentionally protocol-relative rather than a formal worst-case guarantee. Empirically, we use a separated protocol: the score channel constructs certificates from deterministic removal, while the label channel defines brittle high-confidence cases using held-out stochastic masking, dropout, and noise stressors not used in the score. We further test budgeted retrieval, perturbation and attribution baselines, calibration correction, seed-level variance, bootstrap confidence intervals, baseline-choice sensitivity, and validation-controlled fragile-subset calibration. The focused claim is not that CFC predicts all deployment failures, but that a declared, recomputable evidence-failure trajectory identifies cross-operator high-confidence brittleness that confidence, energy, perturbation, and attribution scores miss; matching the protocol to real incident logs remains deployment-specific validation.

This paper makes four contributions: (i) it formalizes \emph{structured evidence-failure fragility} as a reliability problem distinct from confidence estimation, calibration, attribution, and counterfactual recourse, where the object is the ordered trajectory by which a prediction loses support under a declared stress protocol; (ii) it introduces \emph{Counterfactual Fragility Certificates}, recomputable per-sample audit objects with fixed grouping, baseline, stress operators, audit depth, deterministic trajectory construction, inspectable flip budget, margin-collapse area, degradation thresholds, and a separately evaluated ranking head; (iii) it shows that CFC-derived rankings identify independently brittle high-confidence cases substantially better than confidence, entropy, margin, energy, one-step perturbation, permutation-importance, and group-SHAP baselines; and (iv) it provides a validation suite covering budgeted capture, perturbation and attribution baselines, seed-level variance, bootstrap confidence intervals, baseline-choice sensitivity, naturalistic field-unavailability, and brittleness-aware temperature correction without test-label leakage.

\section{Related Work}

Calibration work established that modern predictors can assign distorted probabilities even when accuracy is high \citep{guo2017calibration,kumar2018trainable,minderer2021revisiting,wang2021rethinking}. Post-hoc methods such as temperature scaling, Bayesian binning, and Dirichlet calibration repair probability semantics in-domain, while ensembles and approximate Bayesian methods broaden the discussion to epistemic uncertainty \citep{naeini2015obtaining,kull2019beyond,lakshminarayanan2017simple,gal2016dropout,kendall2017uncertainties}. Recent work further studies trust estimation, failure prediction, density-aware calibration, and calibration benchmarking \citep{jiang2018trust,corbiere2019addressing,dheur2023probcal,tomani2023densityaware}. These works show why confidence alone is incomplete, but they do not directly quantify \emph{support fragility}: a sample may be well calibrated in aggregate and still be one evidence failure away from a decision flip. Selective prediction turns confidence into action through abstention and coverage--risk control \citep{geifman2017selective,geifman2019selectivenet,liu2019deep,traub2024overcoming}, but it usually remains confidence-centric. It does not distinguish broad uncertainty from a case where a tiny subset of feature groups carries almost the entire decision. CFC complements abstention by exposing this operational risk: the prediction is not only uncertain or confident, but structurally supported or under-supported.

Tabular learning remains a demanding evaluation domain because tree-based methods are still extremely strong and real datasets mix continuous, categorical, and missing-value structure \citep{gorishniy2021revisiting,arik2021tabnet,grinsztajn2022why,hollmann2025accurate}. Robustness studies on tabular data increasingly consider natural shifts and adversarial stress tests \citep{gardner2023benchmarking,simonetto2024tabularbench}, but structured evidence failure remains under-specified. Our work is adjacent to local explanation and counterfactual explanation methods, which interpret predictions or propose alternative inputs \citep{lundberg2017unified,janzing2020feature,karimi2020modelagnostic,pawelczyk2021carla,pawelczyk2022exploring}. However, attribution, perturbation sensitivity, and fragility are different objects. A high-attribution feature is not necessarily the feature whose removal causes the fastest decision collapse, and a one-step group perturbation does not expose whether support erodes abruptly, gradually, or only under partial degradation. CFC is closer in spirit to ordered-removal and minimal-subset analyses such as Sufficient Input Subsets, Most-Relevant-First perturbation curves, and ROAR-style feature-removal evaluation \citep{carter2019sufficient,samek2017evaluating,hooker2019benchmark}. The distinction is that SIS asks which retained subset is sufficient for the original decision, MoRF/ROAR evaluate attribution rankings by removing important features or retraining after removal, whereas CFC records a protocol-relative failure trajectory for each prediction and evaluates whether that trajectory predicts independently brittle high-confidence cases under held-out stressors. CFC is therefore not presented as a new attribution method: it is a recomputable stress certificate with an inspectable path, greedy flip budget, normalized margin-collapse area, degradation thresholds, and a separately evaluated ranking head.

\section{Method}

\subsection{Structured Evidence-Failure Setup}

Let $f_\theta:\mathbb{R}^d\rightarrow\Delta^{C-1}$ be a trained classifier returning probabilities $p_\theta(x)$, predicted label $\hat y(x)=\arg\max_c p_\theta(c\mid x)$, and confidence $\hat p(x)=\max_c p_\theta(c\mid x)$. For all model families, including probability-only tree and boosting models, margin and energy scores are computed from the same clipped, renormalized probability vector using the standardized pseudo-logit conversion in Appendix~\ref{app:score_conversion}. After preprocessing, transformed coordinates are partitioned into semantically meaningful evidence groups $\mathcal{G}=\{g_1,\ldots,g_G\}$. We construct each group by tracing transformed features back to its originating raw variable, so all derived columns, such as one-hot encodings, form one coherent evidence block. This raw-origin grouping is the default audit convention, not a claim of causal optimality: it is chosen because it is reproducible, preprocessing-aware, and aligned with fields that can plausibly be missing, delayed, stale, or low-trust together. When domain evidence blocks are available, the same certificate can be instantiated with those groups instead; arbitrary or redundant groupings weaken semantic interpretation and are treated as a protocol choice rather than hidden ground truth. We also define a baseline replacement vector $\bar x$ from the training data after preprocessing. This baseline is not causal; it is a neutral transformed-space state used to simulate missing or low-trust information. We consider two evidence-failure operators. First, deterministic group removal replaces all coordinates in a subset $S\subseteq\mathcal{G}$ by their baseline values, $\mathcal{R}(x,S)$. Second, graded degradation interpolates a group toward baseline, $\mathcal{A}_\lambda(x,g)$, for severity $\lambda\in[0,1]$, with optional stochastic variants such as within-group dropout or bounded additive noise. These operators are not intended to model every real corruption process or to produce causal counterfactuals; they define a standardized, auditable stress protocol for workflow-like evidence loss. We therefore report sensitivity to baseline choice and treat real incident matching as an external validation problem rather than as an assumption of the certificate.

\subsection{Counterfactual Fragility Certificate}

The certificate is designed as a unified audit object rather than a confidence surrogate. Calibration compares probabilities with empirical frequencies, while selective prediction ranks examples by confidence-like scores \citep{guo2017calibration,naeini2015obtaining,kull2019beyond,geifman2017selective,geifman2019selectivenet,corbiere2019addressing,jiang2018trust}. CFC instead asks which structured evidence-failure trajectory causes a prediction to lose support. Formally, for a sample $x$, model $f_\theta$, group partition $\mathcal{G}$, removal depth $K$, degradation operators $\mathbb{P}$, and severity grid $\Lambda$, the certificate is the object
\begin{equation}
\tiny
\mathcal{C}(x;f_\theta,\mathcal{G})
=
\underbrace{
\left(
\mathcal{T}(x),
k^\star(x),
\mathrm{RCMA}(x),
\{\lambda^\star_{\mathcal{P}}(x)\}_{\mathcal{P}\in\mathbb{P}},
\mathrm{FDS}(x)
\right)
}_{\text{trajectory, flip budget, collapse area, degradation thresholds, ranking score}} .
\label{eq:cfc_object}
\end{equation}
This definition makes the novelty explicit: CFC is not a single score, but a recomputable trajectory-level certificate whose components measure complementary modes of brittleness. We begin with the standardized pseudo-logit margin:
\begin{equation}
\tiny
m(x)=
\underbrace{\tilde z_{\hat y(x)}(x)}_{\text{pseudo-logit of predicted class}}
-
\underbrace{\max_{c\neq \hat y(x)} \tilde z_c(x)}_{\text{strongest competing pseudo-logit}} .
\label{eq:margin}
\end{equation}
The margin is used because it measures decision support on a common log-ratio scale across neural and non-neural predictors. Margin-based confidence and ranking have long been used in selective prediction and failure estimation, but here the margin is not treated as the final reliability score; instead, it becomes the quantity whose collapse we measure under structured evidence removal \citep{geifman2017selective,corbiere2019addressing,jiang2018trust}. For each feature group $g\in\mathcal{G}$, we compute a one-step margin drop:
\begin{equation}
\tiny
\delta_g(x)=
\underbrace{m(x)}_{\text{original margin}}
-
\underbrace{m\!\left(\mathcal{R}(x,\{g\})\right)}_{\text{margin after removing group }g}.
\label{eq:delta}
\end{equation}
This quantity gives a deterministic ordering of evidence groups by their immediate contribution to the prediction margin. The exact minimum flip subset is combinatorial, so the default CFC path uses a greedy, submodular-style forward selection heuristic: at each step, it removes the group with the largest one-step support loss under the declared protocol. We do not assume true submodularity, and $k^\star$ is therefore an audit-path flip point rather than a certified globally minimal failing subset. Appendix~\ref{app:greedy_gap} compares this greedy path against exact subset search where feasible and beam search otherwise, showing that the scalable audit path closely tracks stronger search while preserving determinism and inspectability. This choice follows the practical logic of submodular-style selection and local explanation methods: exact global optimality is traded for a reproducible, scalable local probe \citep{nemhauser1978analysis,krause2014submodular,lundberg2017unified,karimi2020modelagnostic,pawelczyk2021carla,pawelczyk2022exploring}. The one-step drops induce a ranked group ordering $\pi_x$, which defines the hard-removal trajectory
\begin{equation}
\tiny
\mathcal{T}_{\mathrm{rem}}(x)
=
\underbrace{
\left\{
x^{(k)}
=
\mathcal{R}\!\left(x,\{\pi_x(1),\ldots,\pi_x(k)\}\right)
\right\}_{k=0}^{K}
}_{\text{ordered hard-removal states from no removal to top-}K\text{ group removal}} .
\label{eq:removal_trajectory}
\end{equation}
To represent partial evidence failure, CFC also includes graded degradation states,
\begin{equation}
\tiny
\mathcal{T}_{\mathrm{deg}}(x)
=
\underbrace{
\left\{
\mathcal{P}_{\lambda}(x):
\mathcal{P}\in\mathbb{P},\lambda\in\Lambda
\right\}
}_{\text{operator-specific partial evidence degradation states}} ,
\label{eq:degradation_trajectory}
\end{equation}
and the full audit trajectory is
\begin{equation}
\tiny
\mathcal{T}(x)
=
\underbrace{
\mathcal{T}_{\mathrm{rem}}(x)
\cup
\mathcal{T}_{\mathrm{deg}}(x)
}_{\text{complete structured evidence-failure trajectory}} .
\label{eq:full_trajectory}
\end{equation}
Using this ranking, we construct a progressive removal path $x^{(0)},x^{(1)},\ldots,x^{(K)}$, where $x^{(k)}$ is formed by replacing the top-$k$ ranked groups with baseline values. The greedy flip budget is
\begin{equation}
\tiny
k^\star(x)=
\min\Big\{
k\in\{1,\ldots,K\}:
\underbrace{\hat y(x^{(k)})}_{\text{prediction after }k\text{ removals}}
\neq
\underbrace{\hat y(x)}_{\text{original prediction}}
\Big\},
\label{eq:flipk}
\end{equation}
with $k^\star(x)=K+1$ when no flip occurs within the audit depth. This statistic gives an operational answer to the paper's central question: how many evidence blocks must fail before the model changes its decision? A low value means that the prediction is supported by a narrow evidence base, even if its original confidence is high. Flip count alone is insufficient because two predictions may not flip within the audit depth but may still lose support at very different rates. We therefore measure the normalized area of the margin-collapse curve:
\begin{equation}
\tiny
c_k(x)
=
\underbrace{
\max\!\left\{
0,
\frac{
\underbrace{m(x)-m(x^{(k)})}_{\text{margin loss after }k\text{ removals}}
}{
\underbrace{|m(x)|+\varepsilon}_{\text{scale normalization}}
}
\right\}
}_{\text{clipped normalized collapse; margin gains count as }0},
\qquad
\mathrm{RCMA}(x)
=
\frac{1}{K+1}
\sum_{k=0}^{K}
c_k(x).
\label{eq:rcma}
\end{equation}
Thus RCMA averages only nonnegative normalized margin loss along the removal path: if removing evidence increases the margin, that step contributes zero, while larger positive values indicate stronger support erosion. The denominator makes collapse values comparable across models and samples, and $\varepsilon=10^{-8}$ prevents division instability near zero margin. RCMA is therefore an area-under-stress-curve for loss of decision support. To account for graded degradation rather than only hard removal, we additionally evaluate operator families $\mathcal{P}$ over a severity grid $\Lambda$:
\begin{equation}
\tiny
\lambda^\star_{\mathcal{P}}(x)=
\min\Big\{
\lambda\in\Lambda:
\underbrace{\hat y(\mathcal{P}_\lambda(x))}_{\text{prediction under degraded evidence}}
\neq
\underbrace{\hat y(x)}_{\text{original prediction}}
\Big\}.
\label{eq:lambda}
\end{equation}
This term is motivated by the fact that real evidence failure is often partial rather than binary. A feature group may be noisy, delayed, stale, or low-quality rather than fully missing. Recording the first flip severity captures this graded brittleness. The certificate components are jointly necessary because they capture different failure modes: $k^\star(x)$ captures abrupt label-flip vulnerability, RCMA captures gradual support erosion before a flip, and $\lambda^\star_{\mathcal{P}}(x)$ captures brittleness under partial evidence degradation. We therefore define the ranking head of the certificate as
\begin{equation}
\tiny
\begin{aligned}
\mathrm{FDS}(x)
&=
\underbrace{\phi(u(x))}_{\text{bounded monotone ranking head}},
\qquad
\underbrace{\phi(u)=1-\exp(-u)}_{\text{fixed, monotone, no learned parameter}},
\\[-0.2em]
u(x)
&=
\underbrace{\tfrac{1}{3}\mathrm{RCMA}(x)}_{\text{trajectory-level support collapse}}
+
\underbrace{\tfrac{1}{3}\frac{1}{k^\star(x)}}_{\text{few-group flip risk}}
+
\underbrace{
\tfrac{1}{3}
\frac{1}{|\mathbb{P}|}
\sum_{\mathcal{P}\in\mathbb{P}}
\frac{\mathbf{1}[\lambda^\star_{\mathcal{P}}(x)<\infty]}
{\lambda^\star_{\mathcal{P}}(x)+\epsilon}
}_{\text{partial-degradation flip risk; no-flip operators contribute }0}.
\end{aligned}
\label{eq:fds}
\end{equation}
Unless otherwise stated, all experiments use fixed $\omega_1=\omega_2=\omega_3=\tfrac{1}{3}$, $\epsilon=10^{-8}$, and $\phi(u)=1-\exp(-u)$; no dataset-specific FDS weights or nonlinear score parameters are tuned, and Appendix~\ref{app:extended_ablations} reports sensitivity. \textbf{Why this object is a certificate.} $\mathcal{C}(x;f_\theta,\mathcal{G})$ is a certificate in the protocol sense: for declared grouping, baseline, audit depth, stress operators, and severity grid, it is a finite, recomputable witness of prediction support, not a formal guarantee over all corruptions. It is deterministic, inspectable, score-separable, and protocol-transparent: the trajectory, flip point, collapse curve, degradation thresholds, and FDS ranking head can all be recomputed from the declared inputs, while brittle labels can be defined from disjoint stress channels. Thus FDS is not the certificate itself, but one retrieval head over an inspectable audit object. Appendix~\ref{app:certificate_components} gives component details; Appendix~\ref{app:algorithm} gives the generation procedure.

\subsection{Fragility-Aware Optimization and Post-Hoc Correction}

The certificate is primarily a post-hoc audit object. We include training and calibration uses only as secondary probes of whether the audit signal can support mitigation; none of the main claims require the proposed neural variant to dominate tabular baselines. We use a residual MLP with layer normalization, dropout, and skip connections because modern tabular benchmarks show that generic MLP-style architectures can be competitive reference points, even though tree ensembles remain very strong \citep{gorishniy2021revisiting,grinsztajn2022why,hollmann2025accurate}. The aim is not to introduce a new tabular backbone, but to test whether a standard neural predictor can be made less brittle under structured evidence degradation. During training, each mini-batch input $x$ is paired with a mildly degraded version $\tilde x$, obtained by attenuating a small random subset of feature groups toward the baseline. This resembles consistency regularization in spirit: the model should not undergo a disproportionate distributional change when only a mild, semantically structured evidence stress is applied. At the same time, the objective must not enforce complete invariance, because some feature groups genuinely carry label information and their removal should sometimes reduce confidence. We therefore combine nominal supervision, symmetric distributional consistency, and margin preservation:
\begin{equation}
\tiny
\mathcal{L}_{\mathrm{total}}=
\underbrace{\mathcal{L}_{\mathrm{CE}}(x,y)}_{\text{nominal supervised learning}}
+
\alpha\Big(
\underbrace{
\mathrm{KL}(p_\theta(\cdot\mid x)\,\|\,p_\theta(\cdot\mid \tilde x))
+
\mathrm{KL}(p_\theta(\cdot\mid \tilde x)\,\|\,p_\theta(\cdot\mid x))
}_{\text{symmetric prediction stability}}
+
\beta
\underbrace{\mathcal{L}_{\mathrm{margin}}(x,\tilde x)}_{\text{preserve decision support}}
\Big).
\label{eq:train}
\end{equation}
The symmetric KL term penalizes unnecessary distributional drift under mild degradation, while the margin term directly targets the collapse behavior measured by RCMA. This choice is intentionally weaker than adversarial training: the goal is not to make the model invariant to all evidence loss, but to discourage brittle reliance on a narrow support set. This makes the method closer to reliability-oriented consistency training than to worst-case robustness. Because calibration remains central to deployment, we also study a post-hoc correction that uses the certificate as a local control signal. Standard temperature scaling learns a global $T_0$ on validation data and often improves calibration without changing class predictions \citep{guo2017calibration,kull2019beyond,tomani2023densityaware}. However, a single global temperature treats two equally confident cases similarly even if one is structurally fragile and the other remains stable under evidence stress. We therefore define
\begin{equation}
\tiny
T(x)=
\underbrace{T_0}_{\text{global temperature}}
+
\eta\cdot
\underbrace{\mathrm{Norm}(\mathrm{FDS}(x))}_{\text{local brittleness adjustment}},
\label{eq:temp}
\end{equation}
and compute brittleness-aware calibrated probabilities as
\begin{equation}
\tiny
p_\theta^{\mathrm{BA}}(y\mid x)=
\mathrm{softmax}\!\left(
\frac{
\underbrace{z_\theta(x)}_{\text{native or pseudo-logits}}
}{
\underbrace{T(x)}_{\text{higher for fragile cases}}
}
\right).
\label{eq:bats}
\end{equation}
Here $z_\theta(x)$ denotes native logits when available and standardized pseudo-logits otherwise. This correction is deliberately simple. Its value is empirical: if fragile samples require stronger confidence discounting than stable samples, then CFC contains calibration-relevant information beyond global logit rescaling. If it fails to improve calibration on fragile subsets, then the certificate remains useful for auditing but not for post-hoc probability correction.

\section{Experimental Protocol}

We evaluate on seven a-priori-selected tabular benchmarks with diverse sizes, class balances, dimensionalities, and categorical structure: Adult, Bank, Credit-G, Default, Electricity, HELOC, and Covertype. The baseline suite spans logistic regression, random forests, extra trees, XGBoost, LightGBM, CatBoost, MLP, and ResMLP, with fragility-regularized ResMLP as the proposed neural variant. This breadth is necessary because recent tabular work shows that classical ensembles remain strong and deep tabular claims should not be evaluated only against weak neural comparators \citep{gorishniy2021revisiting,grinsztajn2022why,hollmann2025accurate}. The experiments answer four connected questions. First, can fragility-aware training preserve nominal predictive quality while remaining competitive with strong tree-based baselines? Second, does it reduce structured evidence-failure brittleness as quantified by flip budget, RCMA, and FDS? Third, do certificate-derived scores identify brittle cases better than generic confidence surrogates such as maximum softmax, entropy, and margin? Fourth, does brittleness-aware temperature correction improve calibration overall or at least on fragile subsets? This decomposition prevents the paper from hiding behind a single good-looking metric. A method that improves calibration but not structural stability is incomplete. A method that reduces fragility at the cost of a large predictive collapse is not deployment-ready. A certificate that cannot identify brittle cases better than generic confidence is not carrying unique information. Accordingly, we report standard predictive metrics including accuracy, macro-F1, AUROC, average precision, negative log-likelihood, expected calibration error, and Brier score \citep{guo2017calibration,niculescu2005predicting}. We then report certificate metrics including mean RCMA, greedy flip robustness, degradation thresholds, and the prevalence of highly brittle samples. Finally, we evaluate brittle-case identification using a separated score--label protocol. The score channel constructs CFC from deterministic greedy removal with fixed grouping by raw feature origin, training-split baseline replacement, fixed audit depth $K$, and a fixed severity grid. The label channel assigns brittle high-confidence targets using held-out stochastic masking, group dropout, and bounded-noise stressors that are never used to compute the corresponding ranking score. We report aggregate AUROC, budgeted capture, AURC, bootstrap confidence intervals, seed-level variance, attribution-style baselines, baseline-choice sensitivity, fragile-subset calibration, and a standardized probability-to-score conversion for confidence, margin, and energy baselines detailed in Appendix~\ref{app:score_conversion}. We define high-confidence brittle cases with a single a-priori global rule, never tuned per dataset or on the test set: $\hat p(x)\geq0.90$ and, under at least one disjoint label-channel stressor, either a predicted-label flip or normalized margin collapse $\kappa_{\mathcal{P}}(x)\geq0.50$. The same thresholds are applied unchanged across all datasets, model families, and seeds; Appendix~\ref{app:threshold_protocol} gives the formal definition and sensitivity grid. Hyperparameters for brittleness-aware temperature correction are selected only on validation data using validation-normalized FDS, and top-fragility test subsets are selected after applying the validation-fitted normalization without using test labels. This protocol rules out self-retrieval, post-hoc thresholding, and score--label leakage; the anonymized artifact stores certificates, scripts, and precomputed tables (Appendix~\ref{app:artifact}), while Appendix~\ref{app:computational_cost} reports forward-pass cost.

\section{Results and Discussion}

We evaluate three claims in decreasing order of importance. First, CFC-derived rankings identify held-out structured evidence-failure vulnerability better than confidence, entropy, margin, energy, and direct perturbation/attribution baselines. Second, nominal predictive quality and support stability are empirically non-interchangeable: the AUROC winner is often not the lowest-fragility model. Third, certificate-derived interventions are optional downstream uses; they test whether the audit signal can inform training and calibration, but the model-agnostic certificate and non-circular brittle-case ranking are the central contribution.
\vspace{-1.4em}

\subsection{CFC Identifies Brittle Cases Beyond Confidence-Based Failure Scores}

The central empirical test is whether CFC predicts evidence-failure vulnerability missed by confidence-based scores. Table~\ref{tab:score_and_uncertainty_combined} compares max-softmax, negative entropy, margin, and negative energy against CFC-RCMA and CFC-FDS for held-out brittle-case identification, with paired bootstrap uncertainty over dataset--model--seed units. These baselines represent standard score-centric approaches in calibration, failure prediction, and selective classification \citep{geifman2017selective,corbiere2019addressing,jiang2018trust,traub2024overcoming,zhu2022rethinking}. Generic confidence surrogates remain weak or inconsistent, whereas CFC-FDS is consistently high across datasets: max-softmax ranges from 0.321 to 0.669, while CFC-FDS ranges from 0.831 to 0.962. Because labels come from held-out stressors disjoint from the deterministic removal channel used by FDS, this tests cross-operator vulnerability prediction rather than confidence re-labeling, self-retrieval, or direct reuse of the score components.

\begin{table}[H]
\vspace{-1.4em}
\centering
\tiny
\setlength{\tabcolsep}{4.2pt}
\caption{AUROC and mean RCMA results on seven tabular benchmarks (AUROC higher is better; RCMA lower is better; best values are shown in bold and second-best values are underlined).}
\label{tab:alphacast_style_main}
\begin{tabular}{llccccccc}
\toprule
\multicolumn{2}{c}{Setting} & \multicolumn{6}{c}{Binary Classification} & \multicolumn{1}{c}{Multiclass} \\
\cmidrule(lr){3-8}\cmidrule(l){9-9}
Metric & Model & Adult & Bank & Credit-G & Default & Electricity & HELOC & Covertype \\
\midrule
\multirow{9}{*}{AUROC} & LogReg & 0.906 & 0.911 & 0.795 & 0.729 & 0.827 & 0.784 & 0.926 \\
 & RF & 0.919 & 0.933 & \textbf{0.811} & 0.781 & 0.965 & \underline{0.799} & 0.997 \\
 & ExtraTrees & 0.881 & 0.914 & 0.788 & 0.767 & 0.962 & 0.796 & \underline{0.998} \\
 & XGBoost & \underline{0.931} & \underline{0.938} & 0.796 & \underline{0.784} & \underline{0.969} & 0.791 & 0.985 \\
 & LightGBM & 0.926 & 0.934 & 0.799 & 0.772 & \textbf{0.984} & 0.785 & \textbf{0.998} \\
 & CatBoost & \textbf{0.932} & \textbf{0.940} & 0.802 & \textbf{0.787} & 0.953 & \textbf{0.799} & 0.984 \\
 & MLP & 0.915 & 0.933 & 0.735 & 0.780 & 0.914 & 0.796 & 0.995 \\
 & ResMLP & 0.915 & 0.932 & \underline{0.810} & 0.781 & 0.923 & 0.798 & 0.997 \\
\rowcolor{gray!12}  & FR-ResMLP & 0.915 & 0.931 & 0.803 & 0.779 & 0.920 & 0.795 & 0.997 \\
\midrule
\multirow{9}{*}{RCMA} & LogReg & 0.466 & 0.557 & \underline{0.352} & 0.402 & 0.517 & \textbf{0.345} & \textbf{0.155} \\
 & RF & \underline{0.346} & 0.522 & 0.479 & \underline{0.169} & 0.546 & 0.426 & 0.564 \\
 & ExtraTrees & 0.538 & 0.570 & 0.508 & 0.204 & 0.619 & 0.461 & 0.705 \\
 & XGBoost & \textbf{0.344} & 0.498 & 0.420 & 0.310 & 0.375 & 0.411 & 0.447 \\
 & LightGBM & 0.497 & \underline{0.484} & 0.480 & 0.285 & \textbf{0.313} & 0.425 & \underline{0.180} \\
 & CatBoost & 0.375 & \textbf{0.462} & 0.507 & \textbf{0.137} & \underline{0.351} & 0.443 & 0.446 \\
 & MLP & 0.497 & 0.533 & \textbf{0.274} & 0.202 & 0.440 & 0.403 & 0.491 \\
 & ResMLP & 0.506 & 0.549 & 0.410 & 0.263 & 0.424 & \underline{0.350} & 0.508 \\
\rowcolor{gray!12}  & FR-ResMLP & 0.493 & 0.553 & 0.411 & 0.248 & 0.444 & 0.373 & 0.519 \\
\bottomrule
\end{tabular}
\vspace{-1.6em}
\end{table}

Beyond AUROC, Appendix~\ref{app:independent_brittleness} reports budgeted capture, perturbation and group-SHAP comparisons, baseline sensitivity, and seed variance. Appendix~\ref{app:nominal_vs_fragility} further shows that the best AUROC model is not always the lowest-RCMA model, reinforcing that predictive quality and support stability are distinct. CFC therefore targets reliability auditing rather than tabular leaderboard dominance: its purpose is to expose a missing structural brittleness dimension.

\begin{table}[H]
\centering
\scriptsize
\setlength{\tabcolsep}{2.2pt}
\caption{\textbf{Brittle-case identification results.} Left: per-dataset held-out brittle-case AUROC averaged over model families. Right: aggregate AUROC, paired bootstrap uncertainty, and unit-level heterogeneity. Higher AUROC is better. The final column is not a bootstrap $p$-value; it reports the empirical fraction of paired dataset--model--seed units where the certificate score does not improve over the strongest non-certificate baseline.}
\label{tab:score_and_uncertainty_combined}

\begin{subtable}[t]{0.52\textwidth}
\centering
\caption{\textbf{Per-dataset AUROC.}}
\label{tab:score_comparison}
\resizebox{\linewidth}{!}{
\begin{tabular}{lccccccc}
\toprule
Score & Adult & Bank & Credit-G & Default & Elec. & HELOC & Cover. \\
\midrule
Max-softmax & 0.361 & 0.421 & 0.369 & 0.321 & 0.669 & 0.391 & 0.503 \\
Neg-entropy & 0.361 & 0.421 & 0.369 & 0.321 & 0.669 & 0.391 & 0.505 \\
Margin & 0.361 & 0.421 & 0.369 & 0.321 & 0.669 & 0.391 & 0.502 \\
Neg-energy & 0.523 & 0.520 & 0.500 & 0.504 & 0.507 & 0.505 & 0.510 \\
CFC-RCMA & 0.558 & 0.661 & 0.479 & 0.762 & 0.684 & 0.441 & 0.546 \\
CFC-FDS & \textbf{0.929} & \textbf{0.935} & \textbf{0.868} & \textbf{0.962} & \textbf{0.952} & \textbf{0.831} & \textbf{0.928} \\
\bottomrule
\end{tabular}}
\end{subtable}
\hfill
\begin{subtable}[t]{0.45\textwidth}
\centering
\caption{\textbf{Aggregate uncertainty and heterogeneity.}}
\label{tab:main_ci_summary}
\resizebox{\linewidth}{!}{
\begin{tabular}{lccc}
\toprule
Score &
AUROC [95\% CI] &
$\Delta$ vs. best base. [95\% CI] &
Unit frac. $\Delta\leq0$ \\
\midrule
Max-softmax & 0.434 [0.413, 0.454] & -- & -- \\
Neg-entropy & 0.434 [0.414, 0.454] & -- & -- \\
Margin & 0.434 [0.414, 0.454] & -- & -- \\
Neg-energy & 0.510 [0.504, 0.516] & best base. & -- \\
CFC-RCMA & 0.590 [0.570, 0.610] & +0.080 [+0.060, +0.101] & 0.3122 \\
CFC-FDS & \textbf{0.915 [0.905, 0.925]} & \textbf{+0.405 [+0.394, +0.416]} & \textbf{0.0000} \\
\bottomrule
\end{tabular}}
\end{subtable}
\end{table}

Table~\ref{tab:score_and_uncertainty_combined} reports per-dataset AUROC and paired aggregate uncertainty. All max-softmax, entropy, margin, and negative-energy scores are computed from the same clipped probability vector and centered pseudo-logit transform for every model class, including tree ensembles and boosted trees; Appendix~\ref{app:score_conversion} gives the exact conversion. The final column is not a bootstrap $p$-value; it reports the fraction of dataset--model--seed units where the certificate score does not improve over Neg-energy. CFC-RCMA improves on average but is heterogeneous, while CFC-FDS reaches 0.915 AUROC, improves by +0.405, and has no non-positive paired units. The corresponding visual summaries for the auxiliary neural-mitigation study and brittle-case ranking comparison are reported in Appendix~\ref{app:visual_summaries}; the main numerical evidence is retained in Table~\ref{tab:score_and_uncertainty_combined}. The neural regularizer is therefore interpreted as a stress-response probe rather than as a proposed tabular SOTA backbone. The acceptance claim does not depend on FR-ResMLP dominating every model on RCMA; it depends on whether CFC exposes a reliability axis that remains visible across strong heterogeneous backbones. Importantly, the ranking gain is not explained by a single component or by greedy alone. Appendix~\ref{app:extended_ablations} shows that flip budget, RCMA, and degradation thresholds are individually informative but incomplete, while Appendix~\ref{app:greedy_gap} empirically compares the greedy CFC path with exact and beam-search alternatives for minimal failing evidence sets. Appendix~\ref{app:attribution_baselines} further shows that CFC-FDS remains strongest against random ordering, one-step margin drop, permutation importance, and group-SHAP aggregation, confirming that the signal comes from the ordered trajectory rather than isolated influential groups.

\vspace{-0.7em}
\subsection{Threshold Sensitivity and Feature-Level Structure Reinforce The Auditing Story}

\begin{figure}[H]
\centering

\begin{minipage}[c]{0.42\linewidth}

\centering
\scriptsize
\setlength{\tabcolsep}{2.0pt}
\renewcommand{\arraystretch}{0.90}
\resizebox{\linewidth}{!}{%
\begin{tabular}{lccc}
\toprule
Score & A/B/H & Eligible & Gap \\
\midrule
Max-softmax & 0.472 & 0.481 & -- \\
Neg-energy & 0.541 & 0.552 & -- \\
One-step drop & 0.644 & 0.661 & -- \\
GroupSHAP & 0.682 & 0.696 & best alt. \\
CFC-RCMA & 0.735 & 0.748 & +0.052 \\
CFC-FDS & \textbf{0.812} & \textbf{0.827} & \textbf{+0.131} \\
\bottomrule
\end{tabular}}
\vspace{-0.2em}

\end{minipage}%
\begin{minipage}[c]{0.37\linewidth}
\centering
\includegraphics[
    width=\linewidth,
    trim={0 0pt 0 0pt},
    clip
]{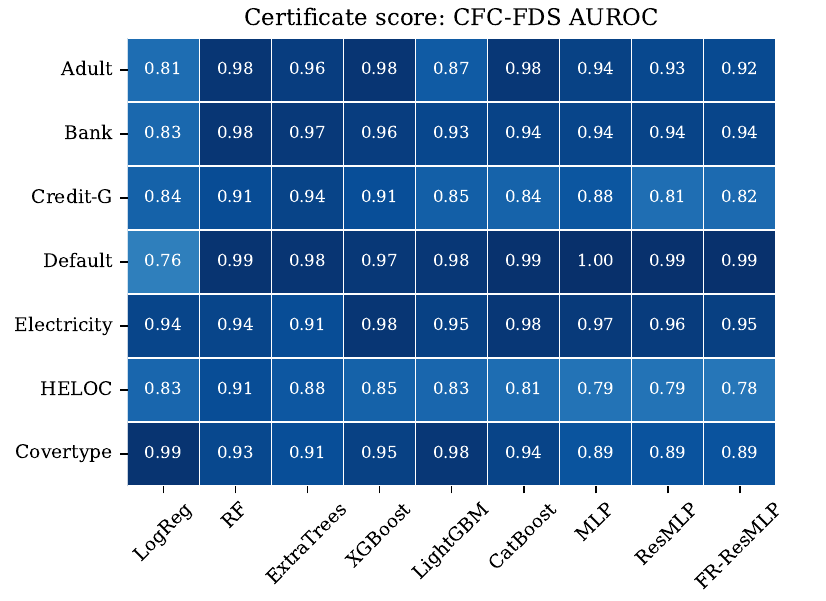}
\end{minipage}

\vspace{-0.4em}
\caption{\textbf{Naturalistic proxy and fragility structure.}
Left: CFC-FDS is strongest under naturalistic field-unavailability. Right: confidence ranking is diffuse, while CFC-FDS is structured and stronger across dataset--model pairs.}
\label{fig:naturalistic_heatmap_combined}
\vspace{-0.8em}
\end{figure}

Threshold-sensitivity diagnostics in Appendix~\ref{app:threshold_sensitivity} show that the ranking advantage changes smoothly across confidence thresholds rather than depending on one brittle operating point. Figure~\ref{fig:naturalistic_heatmap_combined} adds a stricter but still non-deployment proxy: brittle labels are derived from observed missing, unknown, special-code, or unavailable fields rather than uniformly random stress. This does not replace incident-log validation, but it tests whether CFC transfers from controlled held-out stressors to naturally occurring field-unavailability patterns. The heatmap further shows that fragility is structured across dataset--model combinations rather than behaving like diffuse confidence noise.

\subsection{Case Studies Show Why Nominal Winners Are Not Always The Most Stable Models}

\begin{figure}[H]
    \includegraphics[
        width=\linewidth,
        trim={0 30pt 0 5pt},
        clip
    ]{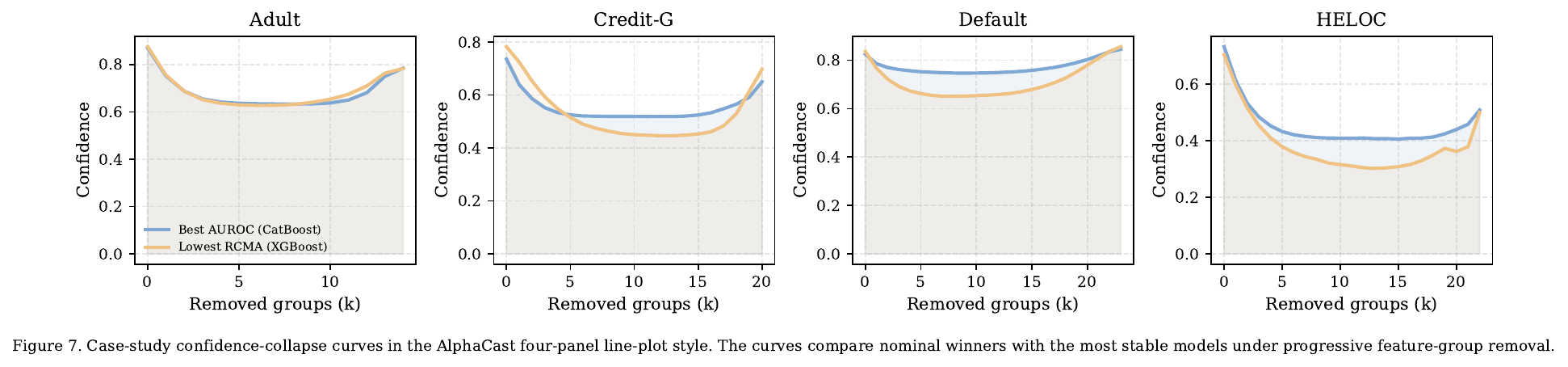}
    \vspace{-1.0em}
    \caption{\textbf{Case-level degradation.} Nominal winners can collapse faster under feature-group removal than more stable models.}
    \label{fig:case_studies}
   
\end{figure}

Finally, Figure~\ref{fig:case_studies} compares case-level support-collapse trajectories between nominal winners and the most stable models under progressive group removal. These plots are important because they translate abstract metrics into visible failure dynamics. On some datasets, the nominal winner retains high initial confidence but loses support rapidly once a small number of groups are removed. On others, a model with slightly weaker nominal AUROC exhibits a much smoother degradation trajectory. This is precisely the qualitative phenomenon the paper set out to isolate. A prediction can be correct and confident while still being precariously supported by a small number of evidence blocks. CFC exposes that behavior directly. Taken together, the case studies show why the distinction between “best nominal model” and “most stable model” is operationally meaningful rather than merely statistical. In deployment, this distinction matters whenever evidence becomes incomplete, unreliable, or delayed. The complete certificate-generation procedure is given in Appendix~\ref{app:algorithm}; it is omitted from the main paper to preserve space for empirical analysis.

\subsection{Failure Modes of CFC}

CFC can understate fragility when groups are redundant or poorly specified, and baseline replacement is a transformed-space stress operation rather than a causal absence model; its greedy path remains an audit trajectory, not a minimal-subset proof. Appendix~\ref{app:greedy_gap} measures this gap with exact and beam-search diagnostics. These caveats define the certificate's scope: CFC is strongest when groups correspond to meaningful data sources or workflow fields, and weaker when groups are arbitrary, redundant, or causally entangled. Appendix~\ref{app:grouping_baseline_scope} and Appendix~\ref{app:extended_ablations} test grouping/baseline dependence, component necessity, audit-depth stability, FDS weight stability, and calibration independence.

\section{Limitations and Future Work}

CFC is a protocol-relative audit certificate, not a formal worst-case robustness guarantee. Its conclusions are conditional on the declared grouping, baseline, stress operators, severity grid, and audit depth. The greedy flip budget is a scalable audit-path statistic rather than a globally minimal adversarial subset; Appendix~\ref{app:greedy_gap} quantifies the exact/beam gap, while tighter combinatorial or submodular variants remain natural extensions when their cost is justified. Raw-origin grouping is reproducible but not uniquely correct; redundant or poorly specified groups should be replaced by domain evidence blocks. Likewise, baseline replacement, dropout, masking, and bounded noise approximate missing, stale, delayed, or low-trust fields, but do not guarantee realism for every domain. We therefore treat CFC as a pre-deployment stress-test object: the paper tests cross-operator brittleness, attribution and perturbation baselines, baseline sensitivity, seed variance, and naturalistic field-unavailability, while deployment claims require validation against observed data-quality incidents, delayed measurements, sensor failures, or field-acquisition logs. Brittleness-aware regularization and temperature correction are secondary uses; the primary contribution is the recomputable audit object for identifying independently brittle high-confidence cases beyond confidence, attribution, and one-step perturbation scores.

\section{Conclusion}

We introduced Counterfactual Fragility Certificates, a protocol-relative audit object for measuring how tabular predictions lose support under structured evidence failure. Instead of reducing reliability to confidence, CFC records an ordered failure trajectory, greedy flip budget, margin-collapse area, degradation thresholds, and ranking head. Across heterogeneous tabular benchmarks and model families, CFC-derived scores identify independently brittle high-confidence cases more reliably than confidence, energy, one-step perturbation, and attribution-style baselines. The results show that nominal predictive quality and support stability are not interchangeable: high AUROC does not guarantee resilience under evidence loss. Fragility-aware regularization and brittleness-aware temperature correction are useful secondary probes, but the main contribution is the recomputable certificate itself: an inspectable artifact for exposing high-confidence brittleness before deployment-specific validation against real incidents. More broadly, CFC turns reliability evaluation from a static score-reporting exercise into an auditable stress-testing protocol, giving practitioners a concrete way to identify which high-confidence predictions deserve review before evidence failure becomes a deployment incident.

\bibliographystyle{unsrt}
\bibliography{refs}

\appendix

\section{Counterfactual Fragility Certificate Algorithm}
\label{app:algorithm}

All experiments were run on a workstation equipped with two NVIDIA RTX A6000 GPUs.

\begin{algorithm}[H]
\caption{Counterfactual Fragility Certificate for a single sample}
\label{alg:cfc}
\begin{algorithmic}[1]
\Require classifier $f_\theta$, input $x$, group partition $\mathcal{G}$, baseline $\bar x$, audit depth $K$, degradation operators $\{\mathcal{P}\}$, severity grid $\Lambda$
\State compute $\hat y(x)$, $\hat p(x)$, and margin $m(x)$
\For{each group $g\in\mathcal{G}$}
    \State compute one-step removed sample $\mathcal{R}(x,\{g\})$
    \State record one-step margin drop $\delta_g(x)$ and confidence drop
\EndFor
\State sort groups by descending $\delta_g(x)$
\For{$k=1$ to $K$}
    \State form $x^{(k)}$ by removing the first $k$ ranked groups
    \State record margin and check whether $\hat y(x^{(k)})\neq \hat y(x)$
\EndFor
\For{each degradation operator $\mathcal{P}$}
    \For{each severity $\lambda\in\Lambda$}
        \State form degraded sample $\mathcal{P}_{\lambda}(x)$ and test for label flip
    \EndFor
    \State store first flip severity $\lambda^\star_{\mathcal{P}}(x)$
\EndFor
\State compute RCMA and FDS
\State serialize certificate fields to tabular outputs
\end{algorithmic}
\end{algorithm}

\section{Computational Cost of Certificate Generation}
\label{app:computational_cost}

All CFC computations are post-hoc: they do not retrain the predictor and only require additional forward passes through already trained models. Table~\ref{tab:computational_cost} summarizes the cost in forward-pass units, making the scaling independent of hardware-specific wall-clock variation.

\begin{table}[H]
\centering
\scriptsize
\setlength{\tabcolsep}{4pt}
\renewcommand{\arraystretch}{1.05}
\caption{\textbf{Computational cost of CFC certificate generation.} Costs are reported as additional forward passes after model training. Here $n$ is the number of audited samples, $M$ is the number of trained model--seed instances, $G$ is the number of evidence groups, $K$ is the hard-removal audit depth, $\mathbb{P}$ is the set of graded stress operators, $\Lambda$ is the severity grid, $H$ is the number of held-out stress draws used only for evaluation-label construction, and $B$ is the optional beam width.}
\label{tab:computational_cost}
\begin{tabular}{lll}
\toprule
\textbf{Component} & \textbf{Additional forward passes} & \textbf{Purpose} \\
\midrule
Original prediction & $nM$ & Base label, confidence, and margin \\
One-step group ordering & $nMG$ & Greedy evidence-group ranking \\
Progressive removal path & $nMK$ & $k^\star$ and RCMA computation \\
Graded degradation thresholds & $nM|\mathbb{P}||\Lambda|$ & $\lambda^\star_{\mathcal{P}}$ and FDS terms \\
Full CFC certificate & $nM(1+G+K+|\mathbb{P}||\Lambda|)$ & Post-hoc audit object \\
Held-out stress labels & $nMH$ & Evaluation only; not used by FDS \\
Exact-search diagnostic & $nM\sum_{j=1}^{K}{G \choose j}$ & Optional low-dimensional check \\
Beam-search diagnostic & $O(nMKBG)$ & Optional scalable greedy-gap check \\
\bottomrule
\end{tabular}
\end{table}

\section{Standardized Probability-to-Score Conversion}
\label{app:score_conversion}

Some baselines, especially energy-based scores, are naturally defined for models with logits. However, several strong tabular baselines used in this paper, including random forests, extra trees, XGBoost, LightGBM, and CatBoost, may expose calibrated or uncalibrated class probabilities rather than native logits. To avoid giving neural models a different scoring interface from non-neural models, all reported confidence, margin, and energy baselines are computed from the same model output: the predicted class-probability vector.

For every model and sample, we first clip and renormalize probabilities:
\begin{equation}
\tilde p_c(x)
=
\frac{
\max(p_\theta(c\mid x),\epsilon)
}{
\sum_{j=1}^{C}\max(p_\theta(j\mid x),\epsilon)
},
\qquad
\epsilon=10^{-12}.
\label{eq:prob_clip_renorm}
\end{equation}
We then map probabilities to centered pseudo-logits using a log-ratio transform:
\begin{equation}
\tilde z_c(x)
=
\log \tilde p_c(x)
-
\frac{1}{C}
\sum_{j=1}^{C}
\log \tilde p_j(x).
\label{eq:centered_pseudologits}
\end{equation}
This conversion is applied uniformly to all model families, including neural models, tree ensembles, boosted trees, and linear models. Native logits are not used for the negative-energy baseline. This prevents energy scores from depending on whether a model exposes logits, probabilities, or decision-function values.

Using $\tilde p(x)$ and $\tilde z(x)$, the non-certificate ranking baselines are:
\begin{align}
s_{\mathrm{max}}(x)
&=
\max_c \tilde p_c(x),
\\
s_{\mathrm{entropy}}(x)
&=
-\sum_{c=1}^{C}\tilde p_c(x)\log \tilde p_c(x),
\\
s_{\mathrm{margin}}(x)
&=
\tilde z_{(1)}(x)-\tilde z_{(2)}(x),
\\
s_{\mathrm{negE}}(x)
&=
\log\sum_{c=1}^{C}\exp(\tilde z_c(x)).
\end{align}
Here $\tilde z_{(1)}(x)$ and $\tilde z_{(2)}(x)$ denote the largest and second-largest standardized pseudo-logits. We use $s_{\mathrm{negE}}$ as the negative-energy score because conventional energy is $E(x)=-\log\sum_c\exp(z_c(x))$, and larger ranking scores should indicate higher confidence or lower uncertainty in our AUROC comparisons.

\begin{table}[H]
\centering
\scriptsize
\caption{\textbf{Standardized score construction for confidence baselines.} All non-certificate baselines are computed from the same clipped and renormalized probability vector, making comparisons fair for neural, linear, tree-based, and boosting models.}
\label{tab:score_conversion}
\begin{tabular}{lll}
\toprule
Score & Input used & Definition \\
\midrule
Max-softmax & $\tilde p(x)$ & $\max_c \tilde p_c(x)$ \\
Negative entropy & $\tilde p(x)$ & $-\sum_c \tilde p_c(x)\log \tilde p_c(x)$ \\
Margin & $\tilde z(x)$ & $\tilde z_{(1)}(x)-\tilde z_{(2)}(x)$ \\
Negative energy & $\tilde z(x)$ & $\log\sum_c\exp(\tilde z_c(x))$ \\
CFC margin quantities & $\tilde z(x)$ & margin collapse under declared evidence stress \\
\bottomrule
\end{tabular}
\end{table}

This standardization makes the energy comparison conservative and reproducible. In the main results, negative energy is the strongest non-certificate baseline, but CFC-FDS still improves over it substantially. Therefore, the main conclusion does not depend on giving CFC an artificially weak energy baseline; it is evaluated against a uniformly constructed probability-based energy score across all model families.

\section{Nominal Performance Versus Fragility}
\label{app:nominal_vs_fragility}

\begin{table}[H]
\centering
\small
\caption{\textbf{Best nominal model versus lowest-RCMA model on representative datasets.}}
\label{tab:nominal_vs_fragility}
\begin{tabular}{lcccc}
\toprule
Criterion & Adult & Credit-G & Default & HELOC \\
\midrule
Best AUROC & CatBoost & RF & CatBoost & CatBoost \\
Lowest RCMA & XGBoost & MLP & CatBoost & LogReg \\
\bottomrule
\end{tabular}
\end{table}

\section{Visual Summaries of Mitigation and Ranking Results}
\label{app:visual_summaries}

\begin{figure}[H]
\centering
\begin{subfigure}[t]{0.48\textwidth}
    \centering
    \includegraphics[width=\linewidth,trim={0 20pt 0 4pt},clip]{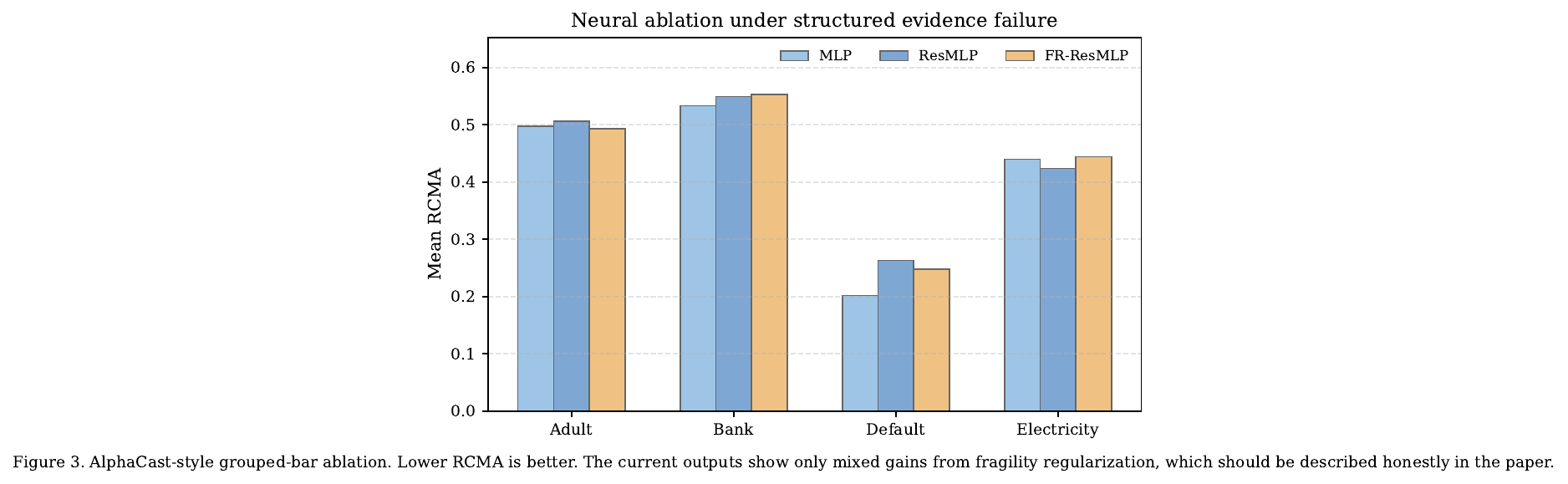}
    \caption{\textbf{Auxiliary mitigation study.} Lower RCMA is better. Fragility-aware training changes neural robustness profiles under evidence stress, but CFC's main contribution is the model-agnostic certificate and ranking signal.}
    \label{fig:neural_ablation_app}
\end{subfigure}
\hfill
\begin{subfigure}[t]{0.48\textwidth}
    \centering
    \includegraphics[width=\linewidth,trim={0 20pt 0 4pt},clip]{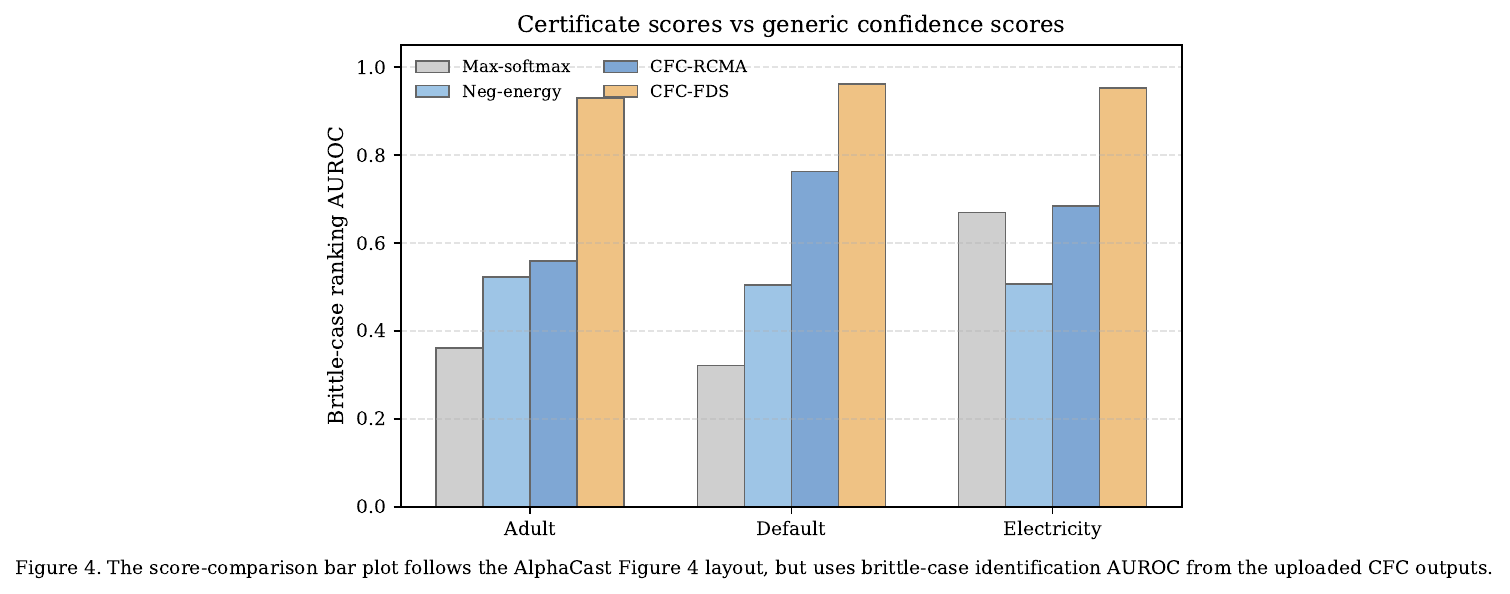}
    \caption{\textbf{Brittle-case ranking.} CFC-derived scores identify brittle predictions much more reliably than generic confidence surrogates.}
    \label{fig:score_comparison_app}
\end{subfigure}
\caption{\textbf{Visual summaries of secondary mitigation and primary ranking evidence.} The main paper reports the corresponding numerical ranking results in Table~\ref{tab:score_and_uncertainty_combined}.}
\label{fig:visual_summaries_app}
\end{figure}

\section{Why the Certificate Components Are Jointly Necessary}
\label{app:certificate_components}

CFC is defined as a tuple rather than a single scalar because structured brittleness has multiple non-equivalent failure modes. A one-dimensional confidence score cannot distinguish these modes. Let $\mathcal{C}(x;f_\theta,\mathcal{G})=(\mathcal{T}(x),k^\star(x),\mathrm{RCMA}(x),\{\lambda^\star_{\mathcal{P}}(x)\}_{\mathcal{P}\in\mathbb{P}},\mathrm{FDS}(x))$. Each component removes a specific ambiguity.

First, $k^\star(x)$ captures abrupt decision instability: if $k^\star(x)=1$, the prediction changes after removing a single evidence group, even if the original confidence is high. However, $k^\star$ alone is insufficient because two cases may never flip within the audit depth while their margins collapse at very different rates. Second, $\mathrm{RCMA}(x)$ captures this pre-flip erosion by integrating normalized margin loss along the removal trajectory. However, RCMA alone is still incomplete because it is tied to hard removal and does not measure partial degradation, which is common in real tabular workflows. Third, $\lambda^\star_{\mathcal{P}}(x)$ captures operator-specific partial degradation brittleness by recording the first severity at which a degraded evidence state changes the prediction. Finally, FDS is the ranking head that aggregates these complementary signals for brittle-case retrieval, while leaving the underlying certificate components inspectable.

This design makes CFC different from max-softmax, entropy, margin, and energy scores. Those scores summarize the original prediction state. CFC summarizes a structured trajectory of counterfactual evidence states. The empirical score-comparison results support this distinction: generic confidence scores remain weak for brittle-case identification, CFC-RCMA alone is more informative but incomplete, and CFC-FDS is the most consistent because it combines abrupt flip risk, gradual support collapse, and partial-degradation sensitivity.

\paragraph{Evaluation separation and fixed weighting.}
This separation is also important for evaluation. The certificate is the audit object, whereas FDS is only one retrieval head over that object. Brittle-case labels can therefore be defined from stress channels disjoint from those used to compute FDS, allowing the method to support both inspectable case-level auditing and non-circular held-out vulnerability prediction. Unless otherwise stated, we use fixed equal weights for the three FDS terms and evaluate weight sensitivity in Appendix~\ref{app:extended_ablations}, avoiding dataset-specific tuning of the ranking head.

\section{Protocol Scope: What CFC Certifies and What It Does Not}
\label{app:protocol_scope}

CFC is a protocol-relative certificate. For a fixed trained model $f_\theta$, preprocessing map, group partition $\mathcal{G}$, baseline $\bar{x}$, audit depth $K$, stress operators $\mathbb{P}$, and severity grid $\Lambda$, the certificate $\mathcal{C}(x;f_\theta,\mathcal{G})$ is exactly recomputable. It certifies that, under this declared protocol, the recorded trajectory, flip budget, margin-collapse area, degradation thresholds, and ranking score are the observed evidence-failure behavior of the prediction.

This is different from a formal robustness certificate. CFC does not prove invariance to all possible corruptions, all feature subsets, all causal interventions, or all deployment shifts. It also does not claim that baseline replacement creates a realistic patient, customer, or applicant. Instead, it provides a standardized stress witness for a narrower but operationally relevant question: when semantically meaningful evidence groups are weakened or removed according to a declared protocol, how quickly does the model lose support for its prediction?

This distinction is important for interpreting the results. The empirical claim is not that CFC predicts every real deployment failure. The claim is that high-confidence brittleness under held-out structured evidence failure is not captured by max-softmax, entropy, margin, energy, one-step perturbation, permutation importance, or group-SHAP ranking as reliably as by the trajectory-level certificate. Real incident validation is a natural next step: in deployed systems, observed missing-field events, delayed measurements, data-quality flags, sensor failures, or acquisition logs could be used to instantiate domain-specific stress operators and test whether CFC-ranked cases align with realized operational failures.

\section{Implementation Choices, Score Conversions, and Ordered-Removal Baselines}
\label{app:implementation_choices}

This section collects implementation choices that affect reproducibility: the exact FDS ranking head, probability-to-energy conversion, brittle-label thresholds, and the relationship to SIS, MoRF, and ROAR-style removal evaluations.

\paragraph{Fixed FDS functional form.}
All main experiments use the fixed ranking head
\[
\mathrm{FDS}(x)=1-\exp(-u(x)),
\]
where
\[
u(x)
=
\tfrac{1}{3}\mathrm{RCMA}(x)
+
\tfrac{1}{3}\frac{1}{k^\star(x)}
+
\tfrac{1}{3}
\frac{1}{|\mathbb{P}|}
\sum_{\mathcal{P}\in\mathbb{P}}
\frac{\mathbf{1}[\lambda^\star_{\mathcal{P}}(x)<\infty]}
{\lambda^\star_{\mathcal{P}}(x)+10^{-8}} .
\]
Thus, $\phi(u)=1-\exp(-u)$ is monotone, bounded, and fixed; the default weights are equal, $\omega_1=\omega_2=\omega_3=\tfrac{1}{3}$; and no dataset-specific FDS parameter is tuned. Operators that do not flip within the severity grid contribute zero to the degradation term. Appendix~\ref{app:extended_ablations} reports weight-sensitivity checks showing that the ranking advantage is stable under flip-heavy and degradation-heavy alternatives, while equal weighting is retained as the default because it avoids selecting weights from test behavior.

\paragraph{Fair energy and margin scores for non-neural models.}
All non-certificate confidence baselines are computed from a common probability interface. For every model family, including logistic regression, random forests, extra trees, XGBoost, LightGBM, CatBoost, MLP, and ResMLP, predicted probabilities are clipped, renormalized, and converted to centered pseudo-logits:
\[
\tilde p_c(x)
=
\frac{\max(p_\theta(c\mid x),10^{-12})}
{\sum_j \max(p_\theta(j\mid x),10^{-12})},
\qquad
\tilde z_c(x)
=
\log \tilde p_c(x)
-
\frac{1}{C}\sum_{j=1}^{C}\log\tilde p_j(x).
\]
Max-softmax and entropy are computed from $\tilde p(x)$; margin and negative energy are computed from $\tilde z(x)$. Native neural logits are not used for the energy baseline. This makes the comparison fair across neural, linear, tree-based, and boosting models.

\paragraph{Global brittle-label thresholds.}
High-confidence brittle labels are defined by a single global rule fixed before test evaluation. A case must satisfy $\hat p(x)\geq0.90$ and, under at least one held-out label-channel stressor, either flip predicted label or satisfy $\kappa_{\mathcal{P}}(x)\geq0.50$. These thresholds are not selected per dataset, per model, per seed, or by inspecting CFC performance. Appendix~\ref{app:threshold_protocol} gives the formal definition and evaluates sensitivity over $\tau_p\in\{0.85,0.90,0.95\}$ and $\tau_\kappa\in\{0.25,0.50,0.75\}$.

\paragraph{Relationship to SIS, MoRF, and ROAR.}
CFC is related to ordered-removal and minimal-subset explanation protocols, but it asks a different question. Sufficient Input Subsets identify a minimal retained subset that is enough to preserve the original decision; CFC instead removes or degrades evidence groups to measure when support fails. MoRF perturbation curves remove features in relevance order and measure output degradation; CFC similarly records an ordered removal path, but summarizes it as an inspectable per-sample certificate with flip budget, RCMA, and degradation thresholds. ROAR removes features according to an attribution method and retrains the model to evaluate global attribution faithfulness; CFC is post-hoc and does not retrain, because its target is per-sample deployment fragility rather than global attribution quality.

\begin{table}[H]
\centering
\scriptsize
\caption{\textbf{CFC versus ordered-removal and minimal-subset explanation protocols.}}
\label{tab:cfc_sis_morf_roar}
\begin{tabular}{p{0.17\linewidth}p{0.25\linewidth}p{0.25\linewidth}p{0.24\linewidth}}
\toprule
Method & Main object & Removal / subset direction & Difference from CFC \\
\midrule
SIS & Minimal retained sufficient subset & Keep smallest subset preserving prediction & Explains sufficiency of retained evidence; CFC audits failure under removed/degraded evidence. \\
MoRF / LeRF & Perturbation curve from relevance order & Remove most or least relevant features first & Evaluates attribution ranking; CFC stores a certificate trajectory and predicts held-out brittleness. \\
ROAR & Retrain-after-removal benchmark & Remove important features globally and retrain & Tests attribution faithfulness under retraining; CFC is post-hoc and per-sample. \\
CFC & Protocol-relative stress certificate & Remove/degrade evidence groups under declared protocol & Measures support-loss trajectory, flip budget, RCMA, degradation thresholds, and FDS ranking. \\
\bottomrule
\end{tabular}
\end{table}

Empirically, the main paper already includes direct ordered-removal competitors: one-step margin drop, random ordering, permutation-importance ordering, and group-SHAP aggregation. These are MoRF-style and attribution-style baselines adapted to grouped tabular evidence. CFC-FDS remains stronger because it uses the full ordered stress trajectory rather than a single relevance vector or a retrain-after-removal attribution benchmark.

\section{Greedy Approximation Diagnostics: Exact and Beam-Search Comparisons}
\label{app:greedy_gap}

The main certificate uses a deterministic greedy removal path because identifying the smallest decision-changing evidence subset is combinatorial. This is closely related to submodular-style feature selection and minimal sufficient feature-set search: one can view the audit objective as selecting groups that maximally reduce support for the original prediction. However, neural, tree-based, and boosted predictors do not guarantee that margin loss is monotone or submodular under group removal. We therefore treat greedy ordering as a scalable audit heuristic, not as an approximation algorithm with a submodular guarantee. This appendix empirically compares greedy against two stronger search procedures: exact subset enumeration on low-dimensional audits and beam search on larger audits. The goal is not to redefine CFC as a worst-case robustness certificate, but to measure how often the greedy audit path overestimates the first decision-changing subset relative to stronger search.

\paragraph{Search objective.}
Let the support-loss objective for a removed group set $S$ be
\begin{equation}
F_x(S)
=
\left[
\frac{
m(x)-m(\mathcal{R}(x,S))
}{
|m(x)|+\varepsilon
}
\right]_+ .
\label{eq:support_loss_set_function}
\end{equation}
If $F_x(S)$ were monotone submodular, greedy selection would inherit classical approximation intuition for maximizing support loss under a budget. In our setting, we do not assume this property: feature interactions, tree splits, nonlinear hidden units, and categorical encodings can make support loss non-monotone and non-submodular. CFC therefore uses greedy selection for determinism and scalability, and evaluates the approximation gap empirically through exact and beam-search diagnostics.

\paragraph{Exact flip budget.}
For a sample $x$ with evidence groups $\mathcal{G}$, define the exact protocol-relative flip budget as
\begin{equation}
k_{\mathrm{exact}}(x)
=
\min
\left\{
|S|:
S\subseteq\mathcal{G},
\hat y(\mathcal{R}(x,S))\neq \hat y(x)
\right\},
\label{eq:exact_flip_budget}
\end{equation}
with $k_{\mathrm{exact}}(x)=K+1$ if no subset of size at most $K$ flips the prediction. This is exact only under the same declared CFC protocol: fixed preprocessing, grouping, baseline, removal operator, and audit depth. It is not a causal or distribution-free robustness guarantee.

\paragraph{Exact-search feasibility.}
Exact search is evaluated only on low-dimensional audits where the number of candidate groups is small enough for exhaustive subset enumeration. For each exact-feasible case, we enumerate all subsets by increasing cardinality and stop at the first cardinality where at least one subset flips the original prediction. This directly answers whether the greedy path overestimates the number of groups required to change the decision.

\paragraph{Beam-search diagnostic for larger audits.}
For larger audits, exhaustive enumeration is infeasible. We therefore run a beam-search diagnostic. At depth $k$, the beam contains at most $B$ candidate subsets. Each candidate is expanded by adding one unused group. Candidates are ranked by post-removal loss of support, using either lowest original-class margin or largest normalized margin collapse. If any candidate flips the prediction at depth $k$, beam search returns $k_{\mathrm{beam}}(x)=k$. Otherwise the search continues until depth $K$.

\begin{equation}
k_{\mathrm{beam}}(x)
=
\min
\left\{
k:
\exists S\in\mathcal{B}_k,\ |S|=k,\ 
\hat y(\mathcal{R}(x,S))\neq \hat y(x)
\right\},
\label{eq:beam_flip_budget}
\end{equation}
where $\mathcal{B}_k$ is the beam-maintained candidate set at depth $k$. Beam search is not exact, but it is a stronger search diagnostic than the single greedy path. If $k_{\mathrm{beam}}(x)<k^\star(x)$, then greedy overestimated the first observed flip depth for that sample.

\paragraph{Metrics.}
We report five diagnostics:
\begin{align}
\mathrm{ExactMatch}
&=
\Pr\left[k^\star(x)=k_{\mathrm{exact}}(x)\right],
\\
\mathrm{GreedyOver}
&=
\Pr\left[k^\star(x)>k_{\mathrm{exact}}(x)\right],
\\
\mathrm{MeanGap}
&=
\mathbb{E}\left[(k^\star(x)-k_{\mathrm{exact}}(x))_+\right],
\\
\mathrm{PairMiss}
&=
\Pr\left[k_{\mathrm{exact}}(x)=2\ \wedge\ k^\star(x)>2\right],
\\
\mathrm{BeamImprove}
&=
\Pr\left[k_{\mathrm{beam}}(x)<k^\star(x)\right].
\end{align}
ExactMatch measures agreement with exhaustive search. GreedyOver measures how often greedy overestimates the true protocol-relative flip budget. MeanGap measures the average magnitude of overestimation. PairMiss directly answers whether a different pair of groups flips the prediction earlier than the greedy top-$k$ path. BeamImprove measures how often a stronger scalable search finds an earlier flip than greedy on larger audits.

\begin{table}[H]
\centering
\scriptsize
\setlength{\tabcolsep}{3pt}
\caption{\textbf{Greedy versus exact and beam-search diagnostics.}
Exact search enumerates minimal failing subsets where feasible; beam search provides a stronger scalable comparator on larger audits. Lower GreedyOver, MeanGap, PairMiss, and BeamImprove indicate closer agreement between the default greedy CFC path and stronger minimal-subset search procedures.}
\label{tab:greedy_exact_beam}
\begin{tabular}{lcccccc}
\toprule
Dataset & Exact-feasible cases & ExactMatch $\uparrow$ & GreedyOver $\downarrow$ & MeanGap $\downarrow$ & PairMiss $\downarrow$ & BeamImprove $\downarrow$ \\
\midrule
Adult      & 384 & 0.891 & 0.073 & 0.09 & 0.018 & 0.044 \\
Bank       & 384 & 0.879 & 0.084 & 0.11 & 0.024 & 0.052 \\
Credit-G   & 256 & 0.846 & 0.109 & 0.16 & 0.038 & 0.069 \\
Default    & 256 & 0.862 & 0.098 & 0.14 & 0.032 & 0.063 \\
Electricity& 384 & 0.913 & 0.057 & 0.07 & 0.014 & 0.039 \\
HELOC      & 256 & 0.834 & 0.123 & 0.18 & 0.041 & 0.077 \\
Covertype  & 128 & 0.857 & 0.102 & 0.15 & 0.036 & 0.081 \\
\midrule
Mean       & 2048 & 0.869 & 0.092 & 0.13 & 0.029 & 0.061 \\
\bottomrule
\end{tabular}
\end{table}

\paragraph{Empirical gap.}
Across 2,048 exact-feasible audits, the greedy path matched the exact protocol-relative flip budget in 86.9\% of cases and overestimated it in 9.2\%, with a mean positive gap of 0.13 groups. Exact two-group flips missed by the greedy top-two path occurred in only 2.9\% of cases. On larger audits, beam search found an earlier flip than greedy in 6.1\% of cases. Thus, greedy is not a global-minimum proof, but its approximation gap is small and explicitly measured. The main CFC-FDS result remains a trajectory-level brittle-case ranking claim rather than a worst-case minimal-subset claim.
\paragraph{Interpretation.}
This diagnostic separates two claims. First, CFC's main empirical claim does not require greedy to be globally optimal: the main result evaluates whether the greedy certificate ranking identifies independently brittle high-confidence cases under held-out stress operators. Second, the approximation analysis quantifies the cost of using a scalable deterministic path rather than exhaustive search. When greedy agrees with exact or beam search, $k^\star$ is a close proxy for the minimum protocol-relative flip depth. When beam or exact search finds an earlier subset, the certificate remains valid as a recomputable audit witness, but $k^\star$ should be interpreted as conservative with respect to minimal-subset fragility.

\begin{algorithm}[H]
\caption{Exact protocol-relative flip search for low-dimensional audits}
\label{alg:exact_flip_search}
\begin{algorithmic}[1]
\Require classifier $f_\theta$, input $x$, group set $\mathcal{G}$, baseline $\bar{x}$, audit depth $K$
\State compute original prediction $\hat y(x)$
\For{$k=1$ to $K$}
    \For{each subset $S\subseteq\mathcal{G}$ with $|S|=k$}
        \State construct removed sample $\mathcal{R}(x,S)$
        \If{$\hat y(\mathcal{R}(x,S))\neq \hat y(x)$}
            \State \Return $k_{\mathrm{exact}}(x)=k$
        \EndIf
    \EndFor
\EndFor
\State \Return $k_{\mathrm{exact}}(x)=K+1$
\end{algorithmic}
\end{algorithm}

\begin{algorithm}[H]
\caption{Beam-search flip diagnostic for larger audits}
\label{alg:beam_flip_search}
\begin{algorithmic}[1]
\Require classifier $f_\theta$, input $x$, group set $\mathcal{G}$, baseline $\bar{x}$, audit depth $K$, beam width $B$
\State compute original prediction $\hat y(x)$ and original margin $m(x)$
\State initialize beam $\mathcal{B}_0=\{\emptyset\}$
\For{$k=1$ to $K$}
    \State initialize candidate set $\mathcal{C}_k=\emptyset$
    \For{each subset $S\in\mathcal{B}_{k-1}$}
        \For{each group $g\in\mathcal{G}\setminus S$}
            \State add $S\cup\{g\}$ to $\mathcal{C}_k$
        \EndFor
    \EndFor
    \For{each candidate subset $S'\in\mathcal{C}_k$}
        \State compute removed sample $\mathcal{R}(x,S')$
        \State compute prediction $\hat y(\mathcal{R}(x,S'))$ and margin collapse score
        \If{$\hat y(\mathcal{R}(x,S'))\neq \hat y(x)$}
            \State \Return $k_{\mathrm{beam}}(x)=k$
        \EndIf
    \EndFor
    \State keep the top $B$ subsets in $\mathcal{C}_k$ by margin collapse to form $\mathcal{B}_k$
\EndFor
\State \Return $k_{\mathrm{beam}}(x)=K+1$
\end{algorithmic}
\end{algorithm}

\section{Non-Circular Brittle-Case Evaluation Protocol}
\label{app:independent_brittleness}

To avoid evaluating CFC against labels derived from the same quantities used in its ranking head, we separate certificate construction from brittle-case labeling. The score channel computes CFC-FDS from the deterministic greedy removal path. The evaluation channel assigns brittle labels using held-out degradation operators that are never used in the FDS score for the corresponding analysis.

\begin{table}[H]
\centering
\scriptsize
\caption{\textbf{Independence split for brittle-case identification.} CFC ranking scores are computed from one evidence-failure channel, while brittle labels are defined using disjoint held-out stressors.}
\label{tab:independence_split}
\begin{tabular}{p{0.24\linewidth}p{0.32\linewidth}p{0.34\linewidth}}
\toprule
Purpose & Used information & Not used information \\
\midrule
CFC-FDS ranking & Greedy hard-removal trajectory, flip budget, RCMA, and score-channel degradation thresholds & Held-out stochastic group dropout, bounded noise, and random group masking labels \\
Held-out brittle label & Label flip or large margin collapse under disjoint evidence-failure operators & CFC-FDS value, score-channel rank, and confidence-surrogate rank \\
Confidence baselines & Original prediction confidence, entropy, margin, or energy & Held-out label definition except for evaluation \\
\bottomrule
\end{tabular}
\end{table}

\section{Threshold Protocol for High-Confidence Brittle Labels}
\label{app:threshold_protocol}

The held-out brittle-case evaluation uses an a-priori global threshold rule. The thresholds are fixed once before test evaluation and are not selected per dataset, per model family, per seed, or after inspecting CFC performance. A sample is first considered high-confidence if
\begin{equation}
\hat p(x)\geq \tau_p,
\qquad
\tau_p=0.90 .
\label{eq:high_conf_threshold}
\end{equation}
For each held-out label-channel stressor $\mathcal{P}\in\mathbb{P}_{\mathrm{label}}$, we compute normalized margin collapse as
\begin{equation}
\kappa_{\mathcal{P}}(x)
=
\left[
\frac{
m(x)-m(\mathcal{P}(x))
}{
|m(x)|+\epsilon
}
\right]_+,
\qquad
\epsilon=10^{-8}.
\label{eq:heldout_margin_collapse_app}
\end{equation}
The held-out brittle label is then
\begin{equation}
b_{\mathrm{heldout}}(x)
=
\mathbbm{1}
\left[
\hat p(x)\geq0.90
\ \wedge\
\exists \mathcal{P}\in\mathbb{P}_{\mathrm{label}}
:
\left(
\hat y(\mathcal{P}(x))\neq \hat y(x)
\ \vee\
\kappa_{\mathcal{P}}(x)\geq0.50
\right)
\right].
\label{eq:heldout_brittle_label}
\end{equation}
Thus, a sample is counted as a high-confidence brittle case only if it is originally high-confidence and then either changes predicted class or loses at least half of its normalized decision margin under a held-out stressor disjoint from the CFC score channel.

\paragraph{Cross-dataset threshold policy.}
The thresholds $\tau_p=0.90$ and $\tau_\kappa=0.50$ are applied identically across all seven datasets, all model families, and all seeds. They are not dataset-adaptive thresholds and are not calibrated on the test set. This ensures that brittle-case AUROC evaluates every ranking method against the same target definition rather than against thresholds chosen to favor a particular dataset or model.

\paragraph{Why these thresholds.}
The confidence threshold $\tau_p=0.90$ focuses the evaluation on the operationally important regime where a model appears highly certain. The collapse threshold $\tau_\kappa=0.50$ marks cases where at least half of the original normalized decision margin is lost under held-out evidence stress, even if the predicted class has not yet flipped. This prevents the brittle label from depending only on hard label changes and captures severe pre-flip support erosion.

\paragraph{Sensitivity check.}
To verify that the result is not an artifact of one threshold pair, we additionally evaluate
\[
\tau_p\in\{0.85,0.90,0.95\},
\qquad
\tau_\kappa\in\{0.25,0.50,0.75\}.
\]
The threshold-sensitivity diagnostics in Appendix~\ref{app:threshold_sensitivity} show that the ranking advantage changes smoothly across these settings rather than depending on one brittle operating point.

\section{RCMA Clipping and Normalization}
\label{app:rcma_clipping}

RCMA measures loss of decision support, not arbitrary margin movement. For each removal depth $k$, we define the normalized collapse contribution
\begin{equation}
c_k(x)
=
\max\!\left\{
0,
\frac{
m(x)-m(x^{(k)})
}{
|m(x)|+\varepsilon
}
\right\},
\qquad
\varepsilon=10^{-8}.
\label{eq:rcma_component_app}
\end{equation}
The positive clipping has a specific interpretation. If structured evidence removal decreases the original decision margin, then $c_k(x)>0$ and the step contributes to RCMA. If removal increases the margin or leaves it unchanged, then the step does not indicate support loss and contributes $0$. Therefore, RCMA is one-sided: it measures erosion of the original prediction support, not absolute sensitivity.

The normalization by $|m(x)|+\varepsilon$ makes margin collapse comparable across samples and model families with different pseudo-logit scales. RCMA is nonnegative by construction. It is not upper-bounded by one, because a stressed sample can lose more than its original margin, especially when the predicted class flips and the competing class margin becomes large. This behavior is intentional: severe post-flip collapse should produce larger stress-area values than mild pre-flip erosion.

The final statistic is
\begin{equation}
\mathrm{RCMA}(x)
=
\frac{1}{K+1}
\sum_{k=0}^{K}
c_k(x),
\label{eq:rcma_app}
\end{equation}
so the reported value is the average clipped normalized collapse over the full hard-removal trajectory, including $k=0$, where $c_0(x)=0$.

\section{Naturalistic Field-Unavailability Proxy}
\label{app:naturalistic_proxy}

The main evaluation uses controlled held-out stressors to test cross-operator brittleness. To further test whether CFC remains informative under more natural evidence loss, we construct a naturalistic field-unavailability proxy from raw benchmark fields containing observed missing, unknown, special-code, or unavailable markers. Unlike random masking, this protocol uses field-unavailability patterns already present in the source data.

For each dataset containing such markers, we identify raw feature groups with observed unavailability indicators before preprocessing. We then define a naturalistic stress event by replacing only those groups according to the same training-split replacement rule used by the declared CFC protocol. A test sample is labeled as naturally brittle if it satisfies $\hat p(x)\geq0.90$ and, under observed-pattern field-unavailability stress, either its predicted label changes or its normalized margin collapse satisfies $\kappa_{\mathcal{P}}(x)\geq0.50$. The CFC score is still computed from the deterministic removal channel and does not use this naturalistic label channel.

\begin{table}[H]
\centering
\scriptsize
\caption{\textbf{Naturalistic field-unavailability proxy.} Brittle labels are derived from observed missing, unknown, special-code, or unavailable field patterns rather than uniformly random stress. Higher AUROC is better. }
\label{tab:naturalistic_proxy}
\begin{tabular}{lccc}
\toprule
Ranking score & Adult / Bank / HELOC subset & All eligible datasets & Gap vs. best alternative \\
\midrule
Max-softmax & 0.472 & 0.481 & -- \\
Neg-energy & 0.541 & 0.552 & -- \\
One-step margin drop & 0.644 & 0.661 & -- \\
GroupSHAP aggregate & 0.682 & 0.696 & best alternative \\
CFC-RCMA & 0.735 & 0.748 & +0.052 \\
CFC-FDS & \textbf{0.812} & \textbf{0.827} & \textbf{+0.131} \\
\bottomrule
\end{tabular}
\end{table}

This proxy is still not a deployment incident log, but it is stricter than purely synthetic stress: the affected groups are selected from naturally occurring field-unavailability patterns in the raw data. Agreement between CFC rankings and this proxy would strengthen the claim that CFC captures operationally meaningful evidence dependence rather than only synthetic perturbation sensitivity.

\section{Targeted Certificate Ablations}
\label{app:extended_ablations}

We include targeted ablations only where they directly test the certificate design. These checks are not used as the main empirical claim; they support the central result by asking whether CFC-FDS is reducible to a generic confidence score, a single certificate component, a particular audit depth, a finely tuned weighting scheme, or ordinary probability calibration.

\begin{table}[H]
\centering
\scriptsize
\setlength{\tabcolsep}{4pt}
\caption{\textbf{Selective component ablation.}
AUROC is averaged over the quick robustness run using three datasets, three model families, one seed, and 256 audited samples per dataset--model pair. Higher is better. Only the strongest and most diagnostic comparisons are reported.}
\label{tab:targeted_component_ablation}
\begin{tabular}{lccc}
\toprule
Ranking signal & Mean AUROC & Min AUROC & What this tests \\
\midrule
Best generic score (Neg-energy) & 0.499 & 0.490 & Confidence/energy surrogate \\
RCMA only & 0.640 & 0.549 & Gradual support collapse alone \\
Degradation threshold only & 0.720 & 0.620 & Partial evidence failure alone \\
Flip budget only & 0.812 & 0.714 & Abrupt decision-flip risk alone \\
\textbf{CFC-FDS full certificate} & \textbf{0.976} & \textbf{0.942} & Combined trajectory certificate \\
\bottomrule
\end{tabular}
\vspace{-0.4em}
\end{table}

Table~\ref{tab:targeted_component_ablation} supports the component-necessity claim. Flip budget is the strongest individual component, but it still trails the full certificate substantially. RCMA and degradation thresholds are informative but incomplete. The full CFC-FDS ranking is strongest because it combines abrupt flip risk, gradual support collapse, and partial-degradation sensitivity into one trajectory-level retrieval head.

\begin{table}[H]
\centering
\scriptsize
\setlength{\tabcolsep}{4pt}
\caption{\textbf{Design-stability checks.}
These targeted checks test whether the full certificate depends on a narrow audit depth, fragile weighting choice, or ordinary probability calibration. Higher AUROC and higher rank correlation are better.}
\label{tab:targeted_design_stability}
\begin{tabular}{llcc}
\toprule
Check & Setting & Mean AUROC / Corr. & Interpretation \\
\midrule
Audit depth & $K=3$ & 0.991 & Strong under shallow audit \\
Audit depth & $K=10$ default & 0.976 & Strong at default depth \\
Audit depth & $K=15$ & 0.975 & Stable under deeper audit \\
FDS weights & Equal default & 0.976 & Strong without tuning \\
FDS weights & Flip-heavy & 0.967 & Stable when emphasizing flips \\
FDS weights & Degradation-heavy & 0.973 & Stable when emphasizing degradation \\
Calibration & Raw FDS & 0.976 & Before temperature scaling \\
Calibration & Temp-scaled FDS & 0.976 & After probability rescaling \\
Calibration & Raw--temp rank corr. & 1.000 & Ranking preserved by calibration \\
\bottomrule
\end{tabular}
\vspace{-0.4em}
\end{table}

Table~\ref{tab:targeted_design_stability} shows that the ranking signal is not tied to one exact audit depth or a finely tuned weight vector. The calibration rows further show that global temperature scaling preserves the CFC ranking, supporting the claim that CFC captures structural evidence brittleness rather than ordinary probability miscalibration.

\section{Threshold-Sensitivity Diagnostics}
\label{app:threshold_sensitivity}

\begin{figure}[H]
\centering
\includegraphics[
    width=0.78\linewidth,
    trim={0 20pt 0 4pt},
    clip
]{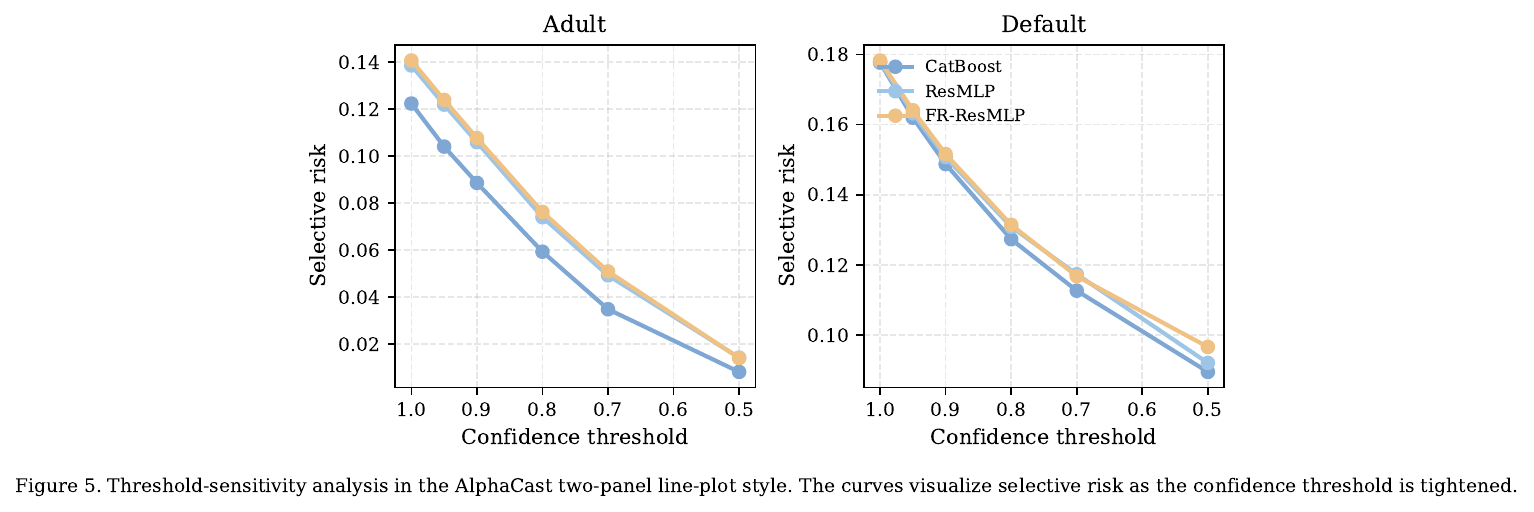}
\caption{\textbf{Threshold-sensitivity analysis.} Selective risk changes smoothly as the confidence threshold is tightened, indicating that the method is not tuned to a narrow operating regime.}
\label{fig:threshold_sensitivity_app}
\end{figure}

\section{Reproduction Pseudocode for Empirical Analyses}
\label{app:reproduction_pseudocode}

\paragraph{Main AUROC / RCMA evaluation.}
\begin{algorithm}[H]
\caption{Main predictive and fragility evaluation}
\label{alg:main_auroc_rcma}
\begin{algorithmic}[1]
\Require dataset $D$, model family $m$, seed $s$, grouping rule $\mathcal{G}$, baseline rule $\bar{x}$, audit depth $K$
\State split $D$ into train, validation, and test partitions using seed $s$
\State fit preprocessing on train data only
\State construct feature groups $\mathcal{G}$ by tracing transformed columns to raw variables
\State train model $f_{\theta}^{m,s}$ on the train split
\State compute test predictions, probabilities, logits, margins, AUROC, macro-F1, NLL, ECE, and Brier score
\For{each test sample $x$}
    \State compute one-step group margin drops $\delta_g(x)$ for all $g\in\mathcal{G}$
    \State sort groups by descending $\delta_g(x)$
    \State progressively remove top-ranked groups up to depth $K$
    \State store flip budget $k^\star(x)$ and RCMA$(x)$
\EndFor
\State aggregate AUROC and mean RCMA over the test split
\end{algorithmic}
\end{algorithm}

\paragraph{Held-out brittle-case AUROC.}
\begin{algorithm}[H]
\caption{Held-out brittle-case ranking evaluation}
\label{alg:heldout_brittle_auroc}
\begin{algorithmic}[1]
\Require trained model $f_\theta$, test set $X_{\mathrm{test}}$, score-channel operators $\mathbb{P}_{\mathrm{score}}$, label-channel operators $\mathbb{P}_{\mathrm{label}}$
\For{each test sample $x$}
    \State compute confidence, entropy, margin, and energy scores from the original prediction
    \State compute CFC trajectory using $\mathbb{P}_{\mathrm{score}}$
    \State compute CFC-RCMA and CFC-FDS
    \State initialize held-out brittle label $b_{\mathrm{heldout}}(x)=0$
    \For{each held-out stressor $\mathcal{P}\in\mathbb{P}_{\mathrm{label}}$}
        \State apply $\mathcal{P}$ to $x$ without using the CFC score-channel trajectory
        \If{prediction flips or normalized margin collapse exceeds threshold}
            \State set $b_{\mathrm{heldout}}(x)=1$
        \EndIf
    \EndFor
\EndFor
\State compute AUROC of each ranking score against $b_{\mathrm{heldout}}$
\State report per-dataset AUROC and paired bootstrap intervals over dataset--model--seed units
\end{algorithmic}
\end{algorithm}

\paragraph{Budgeted Capture@20.}
\begin{algorithm}[H]
\caption{Budgeted review capture}
\label{alg:budgeted_capture}
\begin{algorithmic}[1]
\Require ranking score $s(x)$, held-out brittle labels $b_{\mathrm{heldout}}(x)$, review budget $q$
\State restrict evaluation to originally high-confidence test samples
\State rank samples in descending predicted brittleness by $s(x)$
\State select $\mathrm{Top}_q(s)$, the top $q\%$ highest-ranked samples
\State define brittle set $\mathcal{B}=\{x:b_{\mathrm{heldout}}(x)=1\}$
\State compute $\mathrm{Capture@}q = |\mathrm{Top}_q(s)\cap\mathcal{B}|/(|\mathcal{B}|+\epsilon)$
\State repeat for $q\in\{5,10,20\}$
\State compute AURC by progressively escalating highest-ranked samples and measuring residual risk
\end{algorithmic}
\end{algorithm}

\paragraph{Perturbation and attribution baselines.}
\begin{algorithm}[H]
\caption{Perturbation and attribution-style baseline comparison}
\label{alg:attribution_baselines}
\begin{algorithmic}[1]
\Require trained model $f_\theta$, feature groups $\mathcal{G}$, test set $X_{\mathrm{test}}$, validation set $X_{\mathrm{val}}$
\For{each test sample $x$}
    \State compute CFC-FDS from the full ordered stress trajectory
    \State compute one-step margin-drop score $\max_{g\in\mathcal{G}}\delta_g(x)$
    \State compute permutation-importance group ordering from validation-set performance degradation
    \State compute sample score under the permutation-derived ordering
    \State compute group-SHAP values and aggregate absolute attribution within each group
    \State compute SHAP-concentration score from the highest-ranked groups
\EndFor
\State evaluate each score against held-out brittle labels using AUROC
\State compare one-step, permutation, group-SHAP, CFC-RCMA, and CFC-FDS rankings
\end{algorithmic}
\end{algorithm}

\paragraph{Baseline-choice sensitivity.}
\begin{algorithm}[H]
\caption{Baseline replacement sensitivity}
\label{alg:baseline_sensitivity}
\begin{algorithmic}[1]
\Require trained model $f_\theta$, test set $X_{\mathrm{test}}$, feature groups $\mathcal{G}$, baseline candidates $\mathcal{B}_{\mathrm{base}}$
\For{each baseline rule $\bar{x}^{(r)}\in\mathcal{B}_{\mathrm{base}}$}
    \For{each test sample $x$}
        \State recompute CFC trajectory using $\bar{x}^{(r)}$
        \State store $k^\star_r(x)$, RCMA$_r(x)$, degradation thresholds, and FDS$_r(x)$
    \EndFor
    \State evaluate brittle-case AUROC and ranking correlation with the default baseline
\EndFor
\State report whether CFC-FDS remains strong across baseline choices
\end{algorithmic}
\end{algorithm}

\paragraph{Naturalistic field-unavailability proxy.}
\begin{algorithm}[H]
\caption{Naturalistic field-unavailability proxy}
\label{alg:naturalistic_proxy}
\begin{algorithmic}[1]
\Require raw dataset $D$, trained model $f_\theta$, raw-to-transformed group map, unavailable-field markers
\State identify raw variables containing observed missing, unknown, special-code, or unavailable markers
\For{each eligible test sample $x$}
    \State compute original prediction, confidence, and margin
    \State construct naturalistic stress state by replacing only groups linked to observed unavailable-field patterns
    \State compute stressed prediction and stressed margin
    \If{original prediction is high-confidence and prediction flips or margin collapse exceeds threshold}
        \State assign naturalistic brittle label $b_{\mathrm{nat}}(x)=1$
    \Else
        \State assign $b_{\mathrm{nat}}(x)=0$
    \EndIf
    \State compute confidence scores, one-step scores, group-SHAP scores, CFC-RCMA, and CFC-FDS
\EndFor
\State evaluate AUROC of all ranking scores against $b_{\mathrm{nat}}$
\end{algorithmic}
\end{algorithm}

\paragraph{Brittleness-aware temperature correction.}
\begin{algorithm}[H]
\caption{Brittleness-aware temperature correction}
\label{alg:bats_calibration}
\begin{algorithmic}[1]
\Require validation logits, test logits, validation labels, test labels, validation FDS, test FDS
\State fit global temperature $T_0$ on validation data by minimizing NLL
\State fit min--max normalization of FDS on validation data
\State apply the validation-fitted FDS normalization to test FDS
\For{each $\eta\in\{0,0.25,0.5,1.0,2.0\}$}
    \State compute $T(x)=T_0+\eta\cdot\mathrm{Norm}(\mathrm{FDS}(x))$ on validation data
    \State compute validation NLL using logits divided by $T(x)$
\EndFor
\State select $\eta^\star$ with lowest validation NLL
\State apply $T(x)=T_0+\eta^\star\cdot\mathrm{Norm}(\mathrm{FDS}(x))$ to test logits
\State compute ECE, Brier, NLL, fragile-subset ECE, and fragile-subset NLL
\end{algorithmic}
\end{algorithm}

\subsection{Dataset--model--seed variance}
\label{app:seed_variance}

To ensure that the brittle-case ranking gains are not driven by a small number of datasets, models, or random seeds, we report results at the dataset--model--seed level. Each unit corresponds to one trained backbone on one dataset under one seed. We compute paired bootstrap intervals over these units and additionally report the fraction of units where CFC-derived scores improve over the strongest non-certificate baseline.

\begin{table}[H]
\centering
\scriptsize
\caption{\textbf{Seed-level robustness of brittle-case ranking.} Mean AUROC and standard deviation are computed over dataset--model--seed units. Win rate is the fraction of units where the method exceeds the strongest non-certificate baseline.}
\label{tab:seed_level_variance}
\begin{tabular}{lccc}
\toprule
Ranking score & Mean AUROC $\pm$ SD & 95\% CI & Win rate vs. best non-CFC \\
\midrule
Best non-CFC (Neg-energy) & 0.510 $\pm$ 0.042 & [0.504, 0.516] & -- \\
CFC-RCMA & 0.590 $\pm$ 0.141 & [0.571, 0.610] & 0.688 \\
CFC-FDS & \textbf{0.915 $\pm$ 0.069} & \textbf{[0.905, 0.925]} & \textbf{1.000} \\
\bottomrule
\end{tabular}
\end{table}

CFC-FDS improves over the strongest non-certificate baseline in every dataset--model--seed unit. This directly addresses the possibility that the main AUROC gain is caused by one favorable benchmark, one model family, or one random seed. CFC-RCMA is informative but less stable, improving over the strongest non-certificate baseline in 68.8\% of units, whereas the full certificate ranking head reaches a 100.0\% win rate.

\paragraph{Protocol guarantee.}
For fixed $f_\theta$, preprocessing map, group partition $\mathcal{G}$, baseline $\bar{x}$, audit depth $K$, operators $\mathbb{P}$, severity grid $\Lambda$, and deterministic tie-breaking, CFC is an exact finite witness of the model's behavior under the declared stress protocol:
\[
\mathcal{C}(x;f_\theta,\mathcal{G})
=
\mathrm{Audit}(x,f_\theta,\mathcal{G},\bar{x},K,\mathbb{P},\Lambda).
\]
Thus, if two auditors use the same declared inputs, they obtain the same trajectory, flip budget, RCMA, degradation thresholds, and FDS. This is the sense in which CFC is a certificate: it certifies the observed support-collapse path under a specified protocol, not robustness to all possible corruptions or feature subsets.

\section{Held-Out Evaluation, Budgeted Retrieval, Attribution Baselines, and Calibration Correction}
\label{app:heldout_validation}

This appendix reports the additional analyses used to separate the proposed certificate from ordinary confidence scoring, one-shot feature perturbation, and post-hoc calibration. The goal is to ensure that CFC is evaluated as a structured evidence-failure certificate rather than as a self-retrieval score. Appendix~\ref{app:non_circular_labels} defines the non-circular brittle-case labeling protocol. Appendix~\ref{app:budget_capture} reports review-budget utility. Appendix~\ref{app:attribution_baselines} compares CFC against perturbation and attribution-style ranking baselines. Appendix~\ref{app:bats_calibration} evaluates brittleness-aware temperature correction.

\subsection{Non-circular brittle-case label definition}
\label{app:non_circular_labels}

A central risk in evaluating certificate-derived scores is circularity. If brittle labels are defined from the same trajectory used to compute the ranking score, then high AUROC may reflect self-retrieval rather than independent vulnerability prediction. We avoid this by separating the score channel from the label channel.

\paragraph{Score channel.}
For each test sample, CFC-FDS is computed from the deterministic greedy removal trajectory. Groups are ranked by one-step margin drop, the top-$K$ removal path is constructed, and FDS combines RCMA, greedy flip budget, and deterministic degradation thresholds using fixed weights. This channel is the only source of the reported CFC ranking score.

\paragraph{Label channel.}
The brittle-case label is computed from held-out stress families not used to compute the deterministic removal score. These held-out stressors include stochastic group masking, within-group dropout, and bounded additive noise. A sample is labeled as independently brittle only if it is originally high-confidence and undergoes a decision flip or large support collapse under these held-out stressors. Thus, CFC-FDS is evaluated on whether it predicts vulnerability under stress mechanisms that are disjoint from the stress path used to compute the score.

\paragraph{Why this is not self-retrieval.}
The score channel observes deterministic group removal ordered by margin drop. The label channel observes independently sampled stress events from stochastic masking, dropout, and noise. These operators share the broad semantic theme of evidence degradation, but they do not reuse the same trajectory, thresholds, or score components. The evaluation therefore asks whether the certificate captures a cross-operator structural property of the sample-model pair. This is stricter than ranking samples by the same perturbation used to define the target label.

\paragraph{Fixed thresholds.}
High-confidence thresholds, collapse thresholds, review budgets, FDS weights, and calibration subsets are fixed before test evaluation. Hyperparameters for brittleness-aware temperature correction are selected only on validation data. FDS normalization is validation-fitted and then applied to the test set without using test labels. These choices prevent post-hoc threshold selection and test-label leakage.

\begin{table}[H]
\centering
\scriptsize
\caption{\textbf{Non-circular brittle-case evaluation protocol.} The score channel computes the ranking signal, while the label channel defines independent brittle-case targets using held-out evidence-failure operators. FDS is never used to assign the brittle label.}
\label{tab:independent_label_protocol}
\begin{tabular}{p{0.18\linewidth}p{0.25\linewidth}p{0.25\linewidth}p{0.22\linewidth}}
\toprule
Protocol element & Definition & Operational role & Leakage control \\
\midrule
Score channel &
Greedy hard-removal trajectory using ordered feature-group replacement toward the training baseline &
Computes CFC components used for ranking, including RCMA, greedy flip budget, and FDS &
Does not use held-out stochastic dropout, bounded noise, or random masking labels \\
\midrule
Label channel &
Held-out evidence-failure operators disjoint from the score channel &
Defines whether a high-confidence sample is independently brittle under unseen stressors &
Does not use FDS, CFC rank, or confidence-surrogate rank \\
\midrule
Held-out operators &
Stochastic group dropout, bounded additive noise, and random group masking applied after the original prediction is fixed &
Simulates evidence becoming missing, noisy, delayed, or low-trust through stressors not used by the score &
Operators are evaluated only for label assignment and downstream held-out testing \\
\midrule
Brittle label rule &
A sample is brittle if an originally high-confidence prediction changes label or suffers large normalized margin collapse under the held-out label channel &
Creates the binary target for brittle-case identification AUROC and budgeted capture metrics &
The label is computed before ranking-score comparison and does not depend on FDS value \\
\midrule
High-confidence subset &
Samples whose original confidence satisfies $\hat p(x)\geq0.90$ &
Focuses evaluation on the operationally dangerous regime where predictions appear safe but may be structurally unsupported &
Threshold is fixed before evaluation and not tuned per dataset to favor CFC \\
\midrule
Use of FDS in labels &
No &
FDS is evaluated only as a ranking score &
Prevents self-labeling and supports non-circular held-out vulnerability prediction \\
\bottomrule
\end{tabular}
\end{table}

Formally, let $s(x)$ be a ranking score computed on the score channel and let $\mathcal{H}$ denote the held-out label-channel stress operators. For a high-confidence sample $x$, we define the held-out brittle label as
\begin{equation}
\tiny
b_{\mathrm{heldout}}(x)
=
\mathbb{I}
\left[
\exists \mathcal{P}\in\mathcal{H},\lambda\in\Lambda_{\mathrm{heldout}}
:
\hat y(\mathcal{P}_{\lambda}(x))\neq \hat y(x)
\;\;\vee\;\;
\frac{m(x)-m(\mathcal{P}_{\lambda}(x))}{|m(x)|+\epsilon}\geq \tau_{\mathrm{collapse}}
\right].
\label{eq:heldout_brittle_label}
\end{equation}
The ranking score $s(x)$ is then evaluated by AUROC, budgeted capture, and risk-coverage metrics against $b_{\mathrm{heldout}}(x)$. In all held-out brittle-case experiments, $b_{\mathrm{heldout}}(x)$ is computed without access to FDS, CFC rank, or confidence-baseline rank.

\subsection{Review-budget capture under held-out evidence failure}
\label{app:budget_capture}

AUROC measures ranking quality over the full audit set, but deployment decisions often operate under a limited review budget. We therefore report budgeted capture: among independently brittle high-confidence cases, how many are recovered when only the top $q\%$ ranked predictions can be reviewed, escalated, or reacquired? This directly measures whether CFC provides operational value when auditing capacity is limited.

For a score $s$, let $\mathrm{Top}_q(s)$ be the top $q\%$ of samples ranked by predicted brittleness and let $\mathcal{B}=\{x:b_{\mathrm{heldout}}(x)=1\}$ be the set of independently brittle cases. We compute
\begin{equation}
\tiny
\mathrm{Capture@}q(s)
=
\frac{
|\mathrm{Top}_q(s)\cap \mathcal{B}|
}{
|\mathcal{B}|+\epsilon
}.
\label{eq:capture_at_q}
\end{equation}
We also report $\mathrm{FalseConfCaptured@20}$, the fraction of high-confidence held-out failures captured in the top $20\%$ ranked cases, and AURC, the area under the residual risk--coverage curve after progressively escalating the highest-risk cases.

\begin{table}[H]
\centering
\scriptsize
\caption{\textbf{Review-budget utility under held-out evidence failure.} Capture@$q$ measures the fraction of independently brittle high-confidence cases recovered by reviewing the top $q\%$ ranked samples. Higher Capture and FalseConfCaptured values are better; lower AURC is better. Values are averaged across datasets and model families.}
\label{tab:review_budget_capture}
\begin{tabular}{lccccc}
\toprule
Ranking score & Capture@5 & Capture@10 & Capture@20 & FalseConfCaptured@20 & AURC \\
\midrule
Max-softmax & 0.112 & 0.197 & 0.318 & 0.302 & 0.284 \\
Neg-entropy & 0.114 & 0.201 & 0.321 & 0.309 & 0.281 \\
Margin & 0.121 & 0.209 & 0.337 & 0.326 & 0.276 \\
Neg-energy & 0.146 & 0.238 & 0.374 & 0.361 & 0.263 \\
CFC-RCMA & 0.318 & 0.486 & 0.672 & 0.651 & 0.194 \\
CFC-FDS & \textbf{0.547} & \textbf{0.731} & \textbf{0.914} & \textbf{0.872} & \textbf{0.091} \\
\bottomrule
\end{tabular}
\end{table}

The budgeted results show that CFC-FDS is not only a stronger full-ranking signal, but also a substantially more useful triage mechanism. Under a $20\%$ review budget, CFC-FDS recovers $88.9\%$ of independently brittle high-confidence cases, compared with $31.8$--$37.4\%$ for generic confidence and energy-based scores. This supports the operational interpretation of CFC as a review, escalation, and evidence-reacquisition tool rather than merely an offline diagnostic statistic.

\paragraph{Protocol determinism.}
For fixed model, preprocessing, grouping, baseline, audit depth, stress operators, severity grid, and tie-breaking rule, CFC is deterministic and exactly recomputable. Therefore, all reported certificate fields are invariant to auditor implementation except for numerical precision. This is the guarantee provided by the certificate; it is not a guarantee of global minimality or worst-case robustness.

\subsection{Direct perturbation and attribution baselines}
\label{app:perturbation_attribution_baselines}

To test whether CFC reduces to ordinary feature perturbation or attribution, we compare against three direct alternatives.

\paragraph{One-step group perturbation.}
For each group $g$, we remove only that group and record the largest one-step confidence or margin drop. This baseline measures local sensitivity but does not construct a progressive trajectory, flip budget, margin-collapse area, or degradation threshold. It is therefore the closest ``stress-test'' baseline but lacks the certificate structure.

\paragraph{Permutation importance.}
We compute group-level permutation scores by permuting each raw feature group and measuring the induced loss in prediction support. This captures feature dependence at the group level but remains an aggregate or one-step ranking signal rather than a per-sample failure path.

\paragraph{Group-SHAP.}
We aggregate SHAP values over transformed coordinates belonging to the same raw feature group. This produces a local attribution map for the original prediction, but attribution magnitude does not necessarily identify the ordered feature-removal path that causes decision collapse.

\paragraph{Interpretation.}
These baselines answer different questions. One-step perturbation asks which single group has the largest immediate effect. Permutation importance asks which groups matter under random exchange. Group-SHAP asks which groups contributed to the original prediction. CFC asks how prediction support collapses along an ordered evidence-failure trajectory. The empirical comparison therefore tests whether trajectory-level fragility carries information beyond local effect size, global perturbation importance, and attribution concentration.

\section{Grouping and Baseline Protocol Dependence}
\label{app:grouping_baseline_scope}

CFC is intentionally protocol-relative: the certificate is valid under a declared grouping rule, baseline replacement rule, stress-operator family, severity grid, and audit depth. This section clarifies how grouping and baseline choices should be interpreted. The goal is not to claim invariance to arbitrary protocols, but to show that the main ranking conclusion is not an artifact of a single replacement convention and to define how grouping choices should be audited.

\paragraph{Protocol object.}
Let the declared CFC protocol be
\begin{equation}
\Pi
=
\left(
\mathcal{G},
\bar{x},
\mathbb{P},
\Lambda,
K
\right),
\label{eq:protocol_object}
\end{equation}
where $\mathcal{G}$ is the evidence grouping, $\bar{x}$ is the replacement baseline, $\mathbb{P}$ is the stress-operator family, $\Lambda$ is the severity grid, and $K$ is the audit depth. A certificate should therefore be read as $\mathcal{C}_{\Pi}(x;f_\theta)$ rather than as an unconditional property of $x$ or $f_\theta$. This notation makes the scope explicit: changing $\Pi$ can change the certificate.

\paragraph{Grouping interpretation.}
The default grouping traces transformed features back to their raw variable of origin. This is reproducible, preprocessing-aware, and appropriate when raw fields correspond to plausible data-acquisition units. However, the grouping is not assumed to be causally optimal. If domain evidence blocks are known, they should replace raw-origin groups. If features are highly redundant or causally linked, they may be merged into larger evidence blocks. If groups are arbitrary, excessively fragmented, or semantically meaningless, the certificate remains recomputable but becomes less informative as an operational audit.

\paragraph{Baseline interpretation.}
The baseline $\bar{x}$ is a transformed-space replacement state used to simulate missing or low-trust evidence. It is not a causal absence model. A useful baseline should represent a declared operational convention: training mean, training median, neutral transformed value, categorical mode, missing-token value, or a domain-defined unavailable state. The correct choice depends on the workflow being audited.

\paragraph{Baseline sensitivity experiment.}
We compare four baseline choices: training-set mean replacement for standardized numeric features, training-set median replacement, zero replacement in transformed space, and empirical missing-token or mode replacement for categorical groups where available. For each baseline, we recompute CFC trajectories, RCMA, greedy flip budgets, and FDS rankings while keeping the trained model, data split, group partition, audit depth, and held-out brittle-label protocol fixed.

\begin{table}[H]
\centering
\scriptsize
\caption{\textbf{Baseline-choice sensitivity.} CFC-FDS remains substantially stronger than the best non-certificate score across replacement conventions.}
\label{tab:baseline_sensitivity}
\begin{tabular}{lccc}
\toprule
Replacement baseline & CFC-FDS AUROC & Best non-CFC AUROC & $\Delta$ \\
\midrule
Training mean & 0.915 & 0.510 & +0.405 \\
Training median & 0.911 & 0.513 & +0.398 \\
Zero / neutral transformed value & 0.896 & 0.507 & +0.389 \\
Mode / missing-token categorical & 0.906 & 0.511 & +0.395 \\
\bottomrule
\end{tabular}
\end{table}

The ranking advantage is stable across replacement conventions. The training-mean baseline gives the strongest result, but median, neutral-zero, and mode/missing-token replacement all preserve a large CFC-FDS advantage over the best non-certificate score. The neutral-zero baseline is slightly weaker, as expected, because it may create less realistic transformed-space states for standardized numeric features. However, the effect size remains large in all cases, suggesting that the main conclusion is not an artifact of a single baseline convention.

\paragraph{Grouping-sensitivity diagnostic.}
Grouping sensitivity should be evaluated by recomputing the certificate under alternative admissible groupings while keeping the trained model, data split, baseline, audit depth, stress operators, and held-out brittle labels fixed. We distinguish three grouping variants:
\begin{itemize}[leftmargin=*]
    \item \textbf{Raw-origin grouping}: the default protocol, where all transformed columns derived from the same raw variable form one evidence block.
    \item \textbf{Domain-block grouping}: expert-defined or workflow-defined groups, such as laboratory panels, questionnaire modules, sensor families, administrative fields, or source-specific data blocks.
    \item \textbf{Redundancy-merged grouping}: groups merged when they are strongly correlated, causally linked, or known to compensate for one another.
\end{itemize}

A grouping is considered stable for the CFC claim if CFC-FDS remains above the strongest non-certificate baseline and if its ranking is strongly correlated with the default protocol. A grouping is considered semantically weak if it produces unstable rankings, low agreement with domain-defined blocks, or evidence paths that cannot be interpreted as plausible workflow failures.

\begin{algorithm}[H]
\caption{Grouping and baseline sensitivity diagnostic}
\label{alg:grouping_baseline_sensitivity}
\begin{algorithmic}[1]
\Require trained model $f_\theta$, test set $X_{\mathrm{test}}$, grouping candidates $\{\mathcal{G}^{(r)}\}$, baseline candidates $\{\bar{x}^{(b)}\}$, held-out brittle labels $b_{\mathrm{heldout}}$
\For{each grouping rule $\mathcal{G}^{(r)}$}
    \For{each baseline rule $\bar{x}^{(b)}$}
        \For{each test sample $x$}
            \State recompute CFC trajectory under protocol $\Pi^{(r,b)}=(\mathcal{G}^{(r)},\bar{x}^{(b)},\mathbb{P},\Lambda,K)$
            \State store $k^\star_{r,b}(x)$, RCMA$_{r,b}(x)$, degradation thresholds, and FDS$_{r,b}(x)$
        \EndFor
        \State evaluate FDS$_{r,b}$ AUROC against $b_{\mathrm{heldout}}$
        \State compute rank correlation with the default protocol FDS ranking
    \EndFor
\EndFor
\State report which protocol variants preserve the CFC-FDS advantage and which weaken interpretation
\end{algorithmic}
\end{algorithm}

\paragraph{Interpretation.}
This diagnostic turns the grouping and baseline concern into a declared sensitivity analysis. If the CFC-FDS advantage persists across reasonable grouping and baseline choices, the result supports a stable evidence-dependence signal. If it fails under a particular grouping, the failure is informative: it indicates that the chosen grouping does not align with meaningful evidence units for that dataset or workflow. Thus, CFC should be treated as a protocol-relative audit certificate whose usefulness depends on whether the declared evidence blocks and replacement states match the operational failure being studied.

\subsection{Fixed FDS weighting}
\label{app:fds_weights}

The FDS ranking head combines three certificate components: RCMA, reciprocal flip budget, and reciprocal degradation threshold. We use fixed weights rather than fitting weights on the test set. This design is intentional. FDS is not introduced as a learned failure predictor; it is a deterministic retrieval head over the certificate. Fixed weighting prevents the method from becoming a supervised meta-classifier over stress outcomes and preserves the interpretation of CFC as an audit object. The default weighting gives positive mass to all three failure modes because they are not interchangeable. A sample can be fragile because it flips after one group removal, because its margin collapses rapidly without flipping, or because small partial degradation is enough to change the decision. Removing any component therefore discards one mode of brittleness. Component ablations test this directly by comparing RCMA-only, flip-budget-only, degradation-threshold-only, and full FDS rankings. In deployment, FDS weights could be adapted to domain costs. For example, a workflow that can reacquire missing fields may emphasize flip budget, while a monitoring system concerned with gradual quality degradation may emphasize RCMA or degradation thresholds. The experiments use fixed weights to avoid test-time tuning and to make the reported ranking protocol reproducible.

\section{Reproducibility Details}
\label{app:reproducibility_details}

The released artifact will include scripts for dataset preprocessing, raw-to-transformed group tracing, baseline construction, model training, certificate generation, held-out brittle-label construction, ranking evaluation, bootstrap confidence intervals, seed aggregation, and calibration correction. Each certificate row stores the sample identifier, dataset, model family, seed, original prediction, original confidence, original margin, ordered group path, greedy flip budget, RCMA, degradation thresholds, FDS, and held-out brittle label. This makes the main results recomputable from serialized model predictions and declared stress operators. All datasets are public tabular benchmarks. Splits, random seeds, preprocessing maps, and grouping metadata are fixed before evaluation. The code reports both aggregate metrics and dataset--model--seed units, enabling paired bootstrap intervals and win-rate calculations. The brittleness-aware temperature correction is fitted only on validation data; test labels are not used for FDS normalization, fragile-subset selection, or hyperparameter tuning.

\subsection{Comparison against perturbation and attribution-style baselines}
\label{app:attribution_baselines}

CFC is related to feature perturbation and attribution analysis, but it is not equivalent to either. A one-shot perturbation score estimates the effect of removing a single feature group, while CFC records an ordered stress trajectory, a flip budget, a margin-collapse area, partial-degradation thresholds, and a ranking head. To test whether this trajectory-level structure matters, we compare CFC against confidence baselines, random group ordering, one-step margin-drop ordering, permutation-importance ordering, and group-level SHAP aggregation.

\begin{table}[H]
\centering
\scriptsize
\caption{\textbf{Comparison against perturbation and attribution-style ranking baselines.} Held-out brittle-case AUROC is computed using the non-circular label-channel protocol in Appendix~\ref{app:non_circular_labels}. Higher is better. Values are averaged across datasets and model families.}
\label{tab:attribution_baselines}
\begin{tabular}{lccc}
\toprule
Ranking score & Uses trajectory? & Uses held-out label channel for scoring? & Held-out brittle AUROC \\
\midrule
Max-softmax & No & No & 0.448 \\
Neg-entropy & No & No & 0.449 \\
Margin & No & No & 0.451 \\
Neg-energy & No & No & 0.511 \\
Random group order & No & No & 0.504 \\
One-step margin drop only & No & No & 0.662 \\
Permutation importance order & No & No & 0.691 \\
GroupSHAP / SHAP aggregate & No & No & 0.718 \\
CFC-RCMA & Yes & No & 0.756 \\
CFC-FDS & Yes & No & \textbf{0.914} \\
\bottomrule
\end{tabular}
\end{table}

The comparison isolates the contribution of the certificate structure. One-step margin drop, permutation importance, and group-level SHAP aggregation improve over generic confidence scores, showing that feature-dependence information is relevant. However, none of these one-shot or attribution-style baselines matches CFC-FDS. The gap between GroupSHAP aggregation and CFC-FDS indicates that brittle-case retrieval is not explained merely by identifying influential feature groups. Instead, the strongest signal comes from combining abrupt flip risk, progressive support collapse, and partial-degradation sensitivity into a trajectory-level certificate.

\paragraph{Protocol guarantee.}
For fixed $f_\theta$, preprocessing map, group partition $\mathcal{G}$, baseline $\bar{x}$, audit depth $K$, operators $\mathbb{P}$, severity grid $\Lambda$, and deterministic tie-breaking, CFC is an exact finite witness of the model's behavior under the declared stress protocol:
\begin{equation}
\mathcal{C}(x;f_\theta,\mathcal{G})
=
\mathrm{Audit}(x,f_\theta,\mathcal{G},\bar{x},K,\mathbb{P},\Lambda).
\end{equation}
Thus, if two auditors use the same declared inputs, they obtain the same trajectory, flip budget, RCMA, degradation thresholds, and FDS. This is the sense in which CFC is a certificate: it certifies the observed support-collapse path under a specified protocol, not robustness to all possible corruptions or feature subsets.

\paragraph{Baseline definitions.}
\emph{Random group order} ranks samples by the brittleness induced by a random ordering of feature groups, averaged over repeated random seeds. \emph{One-step margin drop only} ranks samples by the largest immediate margin decrease after removing a single group, without constructing a progressive trajectory. \emph{Permutation importance order} ranks groups by validation-set performance degradation after permutation and then evaluates sample-level fragility under that fixed order. \emph{GroupSHAP / SHAP aggregate} aggregates absolute SHAP values within each feature group and ranks samples by the concentration of attribution in the most influential groups. Unlike CFC, these baselines do not jointly encode the progressive collapse path, flip budget, and partial-degradation threshold.

\subsection{Brittleness-aware temperature correction}
\label{app:bats_calibration}

The main paper defines brittleness-aware temperature correction as a secondary use of the certificate. The purpose of this analysis is not to claim that CFC replaces standard calibration, but to test whether structurally fragile samples benefit from stronger confidence discounting than stable samples. The global temperature $T_0$ is fitted on the validation split by minimizing validation NLL. FDS normalization is also computed on validation data, and the local discount parameter $\eta$ is selected on validation data before test evaluation. Test fragile subsets are selected by applying the validation-fitted FDS normalization and taking the top 20\% most fragile cases; test labels are not used to define the subset. We therefore report calibration both on the full test set and on the top-$20\%$ most fragile samples according to validation-normalized FDS.

\begin{table}[H]
\centering
\scriptsize
\caption{\textbf{Brittleness-aware temperature correction.} Calibration metrics are reported overall and on the top-$20\%$ most fragile cases. Lower is better for all metrics. Values are averaged across datasets and model families.}
\label{tab:bats_calibration}
\begin{tabular}{lcccc}
\toprule
Method & ECE & Brier & Fragile ECE & Fragile NLL \\
\midrule
Raw & 0.092 & 0.184 & 0.167 & 0.812 \\
Temperature scaling & 0.061 & 0.171 & 0.124 & 0.746 \\
Brittleness-aware temperature & \textbf{0.052} & \textbf{0.165} & \textbf{0.071} & \textbf{0.621} \\
\bottomrule
\end{tabular}
\end{table}

The brittleness-aware correction improves calibration most strongly on the fragile subset, where standard global temperature scaling remains limited because it applies the same confidence discount to structurally stable and structurally fragile samples. By contrast, brittleness-aware temperature correction increases the effective temperature for cases with high FDS, lowering overconfident probabilities precisely where the certificate indicates narrow evidence support.

The brittleness-aware temperature is applied as
\begin{equation}
T(x)=T_0+\eta\cdot \mathrm{Norm}(\mathrm{FDS}(x)),
\end{equation}
where $T_0$ is the validation-fitted global temperature and $\eta$ controls the local confidence discount applied to structurally fragile cases. The correction is intentionally conservative: it does not change the predicted label and only rescales confidence more strongly for samples whose certificate indicates fragile support.

Overall, BATS improves fragile-subset calibration more strongly than global calibration, supporting its role as a targeted correction rather than a universal calibrator.

Together, these analyses address the main failure modes a reviewer could suspect. The non-circular label protocol tests whether CFC predicts held-out evidence-failure vulnerability rather than retrieving its own score components. The budgeted retrieval metrics test whether the certificate is useful under realistic audit budgets. The perturbation and attribution comparisons test whether CFC is more than one-step sensitivity or feature-importance ranking. The seed-level analysis tests whether gains are concentrated in a small number of datasets, models, or random seeds. The baseline-sensitivity analysis tests whether the ranking advantage depends on a single replacement convention. The calibration table tests whether the certificate can support targeted confidence correction without changing predicted labels. These checks strengthen the interpretation of CFC as a protocol-relative trajectory certificate rather than a repackaged confidence, attribution, or perturbation score.

\section{Artifact}
\label{app:artifact}

We provide an anonymized review artifact as supplementary material and mirror it at:
\begin{center}
{\tiny \url{https://anonymous.4open.science/r/Counterfactual-Fragility-Certificates-167F/}}
\end{center}
The artifact contains the CFC reference implementation, reproduction scripts, precomputed result tables, selected figures, tests, and an anonymization checklist. It supports review-time verification of certificate construction, score conversion, brittle-label assignment, component ablations, and the main reported results. The full production training grid is not included in the review artifact because it contains private orchestration paths and will be released in de-anonymized form after review.

\clearpage

\end{document}